\documentclass[letterpaper, 10 pt, conference]{ieeeconf}
\IEEEoverridecommandlockouts
\usepackage{graphicx}
\usepackage{amsmath,amssymb}
\usepackage{subcaption}
\usepackage{booktabs}
\usepackage[table]{xcolor}
\usepackage{tikz}
\usetikzlibrary{arrows.meta,positioning,fit,backgrounds,calc,shapes.geometric}
\usepackage{url}

\graphicspath{{figures/}}

\usepackage{algorithm}
\usepackage{algpseudocode}

\usepackage{microtype}
\usepackage{svg}

\usepackage{titlesec}
\titlespacing*{\section} {0pt}{3pt}{2pt}
\titlespacing*{\subsection} {0pt}{3pt}{3pt}

\tikzset{
  block/.style   = {rectangle, rounded corners=3pt, draw=black!70, semithick,
                     align=center, minimum height=11mm, minimum width=20mm,
                     font=\footnotesize, fill=gray!8},
  hw/.style      = {block, fill=blue!8,   draw=blue!55!black},
  srv/.style     = {block, fill=red!7,    draw=red!55!black},
  act/.style     = {block, fill=gray!12,  draw=black!55},
  bmp/.style     = {block, fill=orange!22, draw=orange!75!black},
  sens/.style    = {block, fill=teal!12,  draw=teal!60!black, minimum width=16mm},
  venc/.style    = {block, fill=orange!16, draw=orange!75!black, minimum width=18mm},
  tenc/.style    = {block, fill=red!12,   draw=red!60!black,   minimum width=18mm},
  dec2/.style    = {block, fill=blue!12,  draw=blue!60!black,  minimum width=18mm},
  prompt/.style  = {rectangle, rounded corners=2pt, draw=red!50!black, semithick,
                     fill=red!5, align=center, font=\scriptsize, inner sep=3pt},
  cov/.style     = {block, fill=violet!10, draw=violet!60!black,
                     minimum width=18mm, minimum height=10mm},
  dec/.style     = {diamond, aspect=1.7, draw=black!70, semithick, align=center,
                     font=\scriptsize, fill=yellow!22, inner sep=1pt},
  fsq/.style     = {rectangle, draw=black!55, minimum size=3.1mm, inner sep=0pt},
  img/.style     = {inner sep=0pt, draw=black!55, line width=0.4pt},
  arr/.style     = {-{Latex[length=2.2mm]}, semithick, black!75},
  lab/.style     = {font=\scriptsize\itshape, fill=white, inner sep=1.2pt},
  glab/.style    = {font=\scriptsize\bfseries}
}

\title{\LARGE \bf 
On the Design, Control, and Experimental Evaluation of a Soft, VLM-Driven Reconfigurable Shock-Absorbing Exoskeleton for UAVs}

 \author{Anna Strikou, Mario A.V. Saucedo, Jakub Haluska, and George Nikolakopoulos}

\begin{document}
\long\def\/*#1*/{}
\maketitle
\thispagestyle{empty}
\pagestyle{empty}

\begin{abstract}
Aerial robots are increasingly deployed for inspection, surveillance, mapping, environmental monitoring, and search-and-rescue in complex, unstructured environments, where unplanned collisions can lead to significant damage or mission failure. 
This paper presents the design, control, and experimental evaluation of a soft, reconfigurable shock-absorbing exoskeleton for a quadrotor. Unlike conventional rigid protective structures, the proposed exoskeleton transitions between a compact \emph{folded} configuration that preserves nominal flight performance and a \emph{deployed} configuration that surrounds the vehicle and absorbs impact energy, providing protection only when required. 
To enable autonomous operation, a vision--language model (CLIPSeg) running as an on-demand perception service processes a single onboard RGB camera stream to assess the surroundings and trigger threshold-based deployment of the protective structure when necessary. The finalized design is manufactured and integrated onto a 5-inch quadrotor platform, and 
its performance is then evaluated through controlled collision flight tests against a static obstacle. 
Across matched-velocity impacts, deploying the exoskeleton reduces peak impact force by up to $79\%$, lowers peak acceleration correspondingly, and lengthens the impact duration by $1.4$--$1.9\times$, yielding smoother deceleration and reduced mechanical loads on the airframe. 
\end{abstract}

\section{INTRODUCTION}
Unmanned aerial vehicles (UAVs) have become integral components of modern robotic systems, enabling a wide range of applications including inspection~\cite{UAVinspections}, mapping~\cite{UAVmapping},
environmental monitoring~\cite{UAVenvironment}, and emergency response~\cite{UAVemergencies}, particularly in environments where human access is limited or hazardous. However, operation in cluttered, unstructured, and human-shared spaces exposes these platforms to unexpected contacts and collisions that can compromise system integrity or lead to mission failure. As guaranteeing strictly collision-free trajectories is often impractical in real-world scenarios, aerial robots are increasingly required to tolerate minor, controlled impacts and recover gracefully from unforeseen interactions. 

A complementary line of research focuses on redesigning robot morphology to
enhance safety by reducing reliance on rigid, stiff, and potentially hazardous
structures through the integration of compliant materials. This approach has
driven the rapid development of soft robotics~\cite{softRobots}, encompassing
flexible-joint robots, flexible-link robots~\cite{FlexLinksArms}, and fully soft robotic systems~\cite{softElementsRobots}. When applied to aerial platforms,
compliance offers the ability to absorb impact energy, reduce peak contact
forces, and improve resilience during unintended interactions. However, fully
soft aerial robots remain difficult to realize, as the strong coupling between
structural flexibility and flight dynamics can compromise controllability,
introduce unwanted oscillations, and ultimately hinder stable flight
performance.

\begin{figure}[t]
\centering
\includegraphics[width=0.99\linewidth]{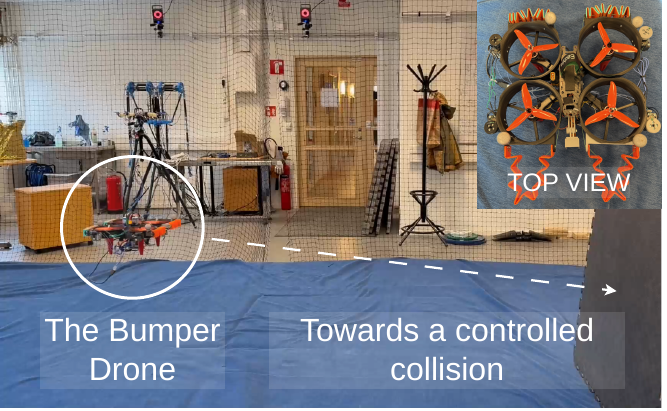}
    \captionsetup{belowskip=-6mm}
\caption{Concept figure of the experimental setup. The Bumper Drone during a controlled collision trial. The UAV is guided toward a fixed obstacle at controlled speed within a motion-capture arena, with its flight behavior tracked throughout. 
Inset: top view of the UAV equipped with the protective exoskeleton, which deploys when approaching the obstacle at unsafe range.}
\label{fig:conceptFigure}
\end{figure}

Approaches to collision resilience in aerial robots broadly fall into three
groups: enclosing the vehicle with external protective structures, building the airframe from soft or morphing materials that absorb energy before failure, and combinations of the two. \textbf{Sensorized and origami structures.} SCoReR~\cite{softProtectiveStructure}
uses a folded PET origami body with compliant bumpers and soft sensors that
detect obstacles \emph{after} contact, focusing on post-collision sensing rather than active protection. An origami/kirigami PEEK structure with shape-memory-alloy (SMA) actuation~\cite{origami} reversibly changes shape to absorb impact energy,
but the protective envelope is considerably larger than the drone, penalizing
aerodynamics and sensor integration. \textbf{Tensegrity shells.} Spherical and icosahedron tensegrity structures~\cite{collisionResiliantStudyForTensegrityStructures,CollisionResilientUAVtansegrityStructure} suspend rigid rods in a tension network and let the vehicle bounce off obstacles; however, several of these studies validate the shell without a fully operational drone
inside. \textbf{Soft and dual-stiffness frames.} Cognifly~\cite{WhenBeingSoftMakesYouTough}
mixes soft and rigid parts in a sub-250\,g quadcopter but is offered only as a
complete, small-scale integrated solution. An insect-inspired dual-stiffness
frame~\cite{InsectInspiredMechanicalResilience} switches between rigid and soft states via magnetic joints, combining stable flight with compliance under high loads. Fully soft, shape-morphing pneumatic platforms~\cite{softPneumaticDroneLTU} demonstrate in-flight morphing but report significant stability issues due to overall flexibility.

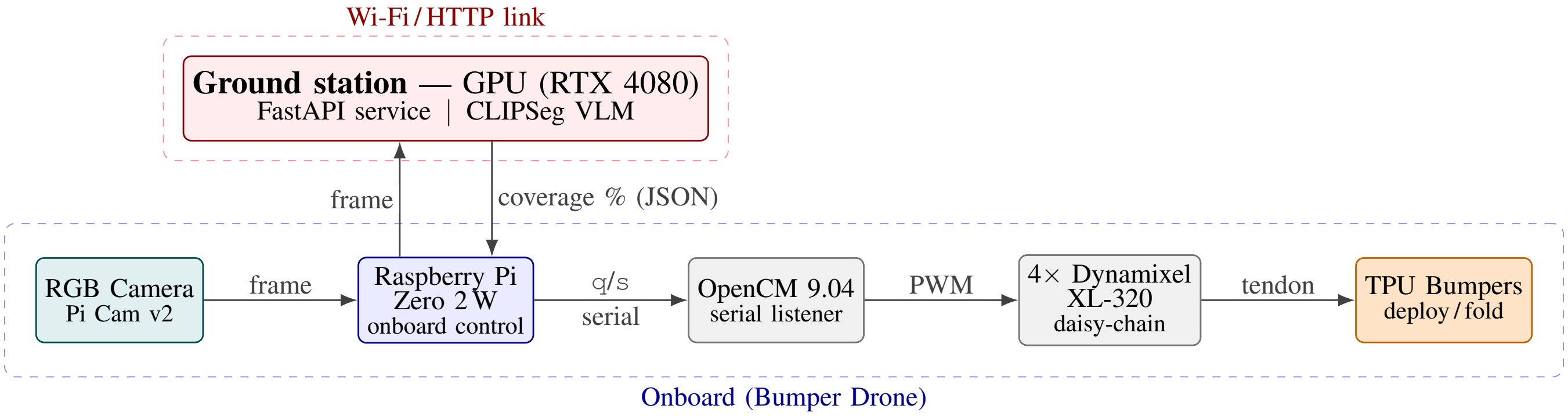
\begin{figure*}[htbp]
\centering
\resizebox{\textwidth}{!}{%
\begin{tikzpicture}[node distance=10mm and 20mm]
\node[sens] (cam) {\normalsize RGB Camera\\\small Pi Cam v2};
\node[hw,  right=of cam]    (rpi)    {\normalsize Raspberry Pi\\\normalsize Zero~2\,W\\\small onboard control};
\node[act, right=of rpi]    (opencm) {\normalsize OpenCM~9.04\\\small serial listener};
\node[act, right=of opencm] (servos) {\normalsize 4$\times$ Dynamixel\\\normalsize XL-320\\\small daisy-chain};
\node[bmp, right=of servos] (bumper) {\normalsize TPU Bumpers\\\small deploy\,/\,fold};
\node[srv, above=15mm of rpi, minimum width=54mm] (server)
     {\large\textbf{Ground station} — GPU (RTX~4080)\\\normalsize FastAPI service \,$\vert$\, CLIPSeg VLM};

\draw[arr] (cam)    -- node[lab,above,font=\normalsize,fill=white,inner sep=2pt]{frame} (rpi);
\draw[arr] (rpi)    -- node[lab,above,font=\normalsize,fill=white,inner sep=2pt]{\texttt{q}/\texttt{s}} node[lab,below,font=\normalsize,fill=white,inner sep=2pt]{serial} (opencm);
\draw[arr] (opencm) -- node[lab,above,font=\normalsize,fill=white,inner sep=2pt]{PWM} (servos);
\draw[arr] (servos) -- node[lab,above,font=\normalsize,fill=white,inner sep=2pt]{tendon} (bumper);
\draw[arr] ([xshift=-6mm]rpi.north) -- node[lab,left,font=\normalsize,fill=white,inner sep=2pt]{frame} ([xshift=-6mm]server.south);
\draw[arr] ([xshift=6mm]server.south) -- node[lab,right,font=\normalsize,fill=white,inner sep=2pt]{coverage \% (JSON)} ([xshift=6mm]rpi.north);

\begin{scope}[on background layer]
\node[draw=blue!45, dashed, rounded corners, inner sep=4mm,
      fit=(cam)(rpi)(opencm)(servos)(bumper),
      label={[glab,blue!55!black,font=\normalsize]below:Onboard (Bumper Drone)}] {};
\node[draw=red!45, dashed, rounded corners, inner sep=2.5mm, fit=(server),
      label={[glab,red!55!black,font=\normalsize]above:Wi-Fi\,/\,HTTP link}] {};
\end{scope}
\end{tikzpicture}%
}
    \captionsetup{belowskip=-4mm}
\caption{System architecture of the Bumper Drone. A single onboard chain streams
camera frames to the Raspberry Pi, which offloads inference to a GPU ground
station running CLIPSeg over Wi-Fi and receives back per-prompt coverage
percentages. Thresholding these values yields a deploy (\texttt{q}) or fold
(\texttt{s}) command that propagates through the OpenCM board and the four
daisy-chained Dynamixel servos to tension the TPU bumpers.}
\label{fig:arch}
\end{figure*}

A recurring limitation across these works is that protection is either
always-on, and therefore aerodynamically costly and bulky, or tightly bound to
a specific small platform. In contrast, the proposed exoskeleton keeps a standard rigid airframe fully intact, adds compliance only around the propellers and motors, and deploys the soft envelope \emph{on demand} using semantic perception. Overall, table~\ref{tab:sota} summarizes the comparison. 
\begin{table}[htbp]
\caption{State-of-the-art reconfigurable/protective structures for UAVs.}
\label{tab:sota}
\centering\renewcommand{\arraystretch}{1.15}
\footnotesize
\begin{tabular}{p{2.1cm} p{5.4cm}}
\toprule
\textbf{Structure} & \textbf{Main limitation} \\
\midrule
SCoReR~\cite{softProtectiveStructure} & Soft sensing focused on \emph{post}-collision detection. \\
Tensegrity~\cite{collisionResiliantStudyForTensegrityStructures} & Validated without an operational drone inside. \\
Cognifly~\cite{WhenBeingSoftMakesYouTough} & Integrated solution limited to small-scale robots. \\
Insect-inspired~\cite{InsectInspiredMechanicalResilience} & Mostly soft. Reduced flight stability. \\
Soft drone~\cite{softPneumaticDroneLTU} & Fully soft. Notable stability issues. \\
Origami/SMA~\cite{origami} & Protective structure larger than the drone. Aerodynamics issues. \\
\midrule
\textbf{This work} & Rigid airframe preserved. Soft envelope deployed \emph{on demand} via a VLM. \\
\bottomrule
\end{tabular}
\end{table}

This trade-off motivates the development of \emph{reconfigurable} structures,
where a fundamentally rigid and stable platform can selectively adopt the
advantages of soft robotics when required. In open environments, a rigid
configuration is maintained to preserve flight stability and efficiency,
whereas operation near humans, infrastructure, or obstacles benefits from
increased compliance to enhance collision tolerance and reduce impact severity.
Building upon this concept, we introduce the \emph{Bumper Drone}: a conventional
5-inch quadrotor augmented with a soft, reconfigurable protective exoskeleton that
deploys only when the onboard VLM-based perception system predicts an elevated risk of
collision. The overall system architecture is presented in Fig.~\ref{fig:arch}, while the main contributions of this work are:
\begin{itemize}
\item The design and development of a novel soft, origami-inspired TPU exoskeleton integrated onto a fully functional rigid quadrotor, capable of morphing between a compact folded configuration and a deployed shock-absorbing state, based on a tendon-driven servo actuation system distributed across four aerodynamic shrouds.
\item An autonomous, language-driven deployment framework that leverages an
open-vocabulary vision--language model to estimate collision risk from
a single RGB camera and trigger protective deployment.
\item An extensive experimental evaluation using motion-capture ground truth to quantify the exoskeleton's effectiveness in reducing impact force and acceleration, while increasing contact duration during collision events.
\end{itemize}

\section{SOFT SYSTEM DESIGN AND METHODOLOGY}
\subsection{Platform Overview}
\begin{figure}[htbp]
\centering
\includegraphics[width=0.99\linewidth]{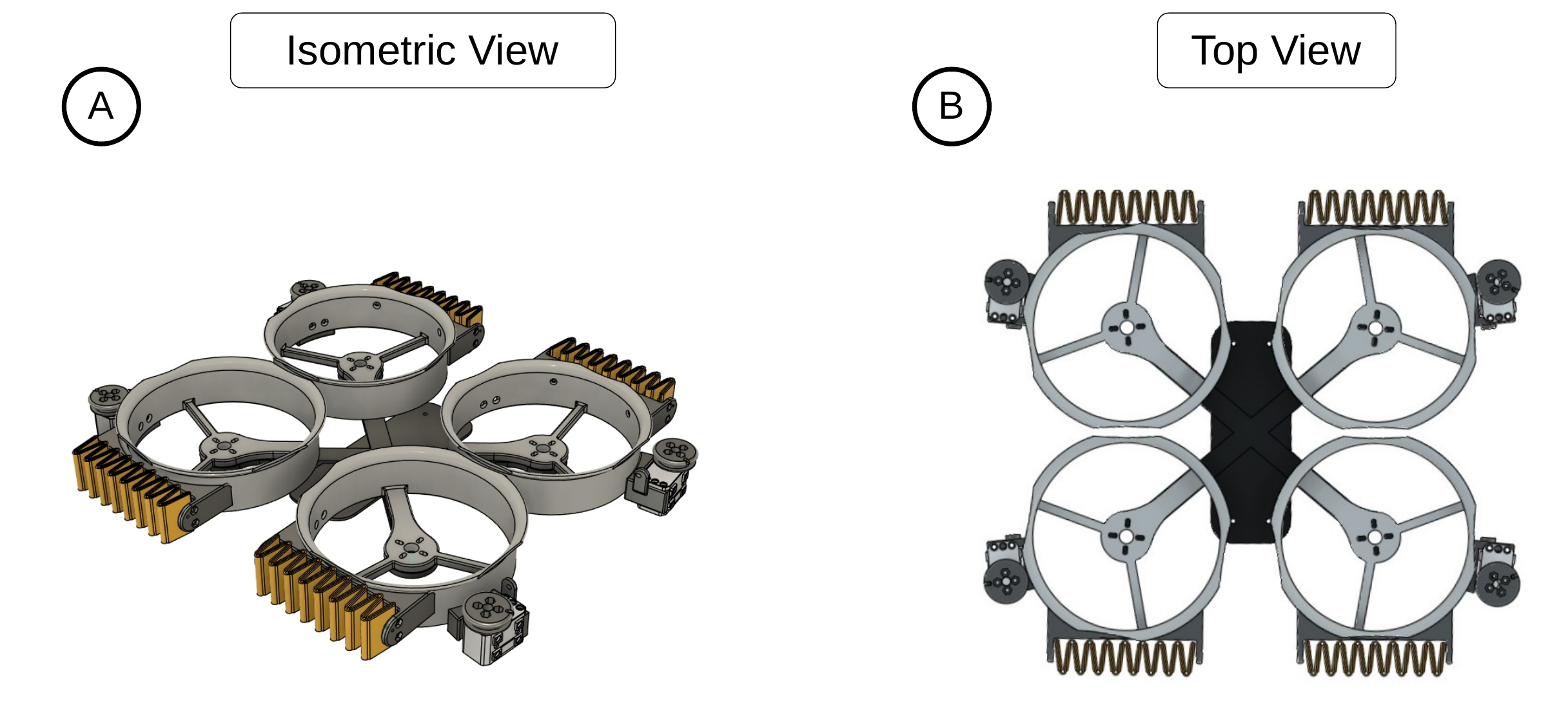}
    \captionsetup{belowskip=-4mm}
\caption{The Bumper Drone. (A) Isometric view; (B) top view. Rigid parts are
shown in gray and the soft, reconfigurable bumpers in yellow.}
\label{fig:overview}
\end{figure}

The Bumper Drone is built on a 5-inch quadrotor in ``X'' configuration with a
rigid carbon-fiber frame, which preserves flight stability and control
authority. A reconfigurable exoskeleton surrounds the frame and the functional
components (motors and propellers) so that they are protected on impact.
Fig.~\ref{fig:overview} shows the assembly: rigid parts (motor shrouds,
brackets, pulleys) are shown in gray and the soft reconfigurable bumpers in
yellow. Four servomotors, one per shroud, actuate the four mechanically
decoupled bumpers. The footprint is $325\times350$~mm with the bumpers folded,
and the overall height reaches $120$~mm once electronics and landing legs are
mounted.

\subsection{Soft Reconfigurable Structure}
\begin{figure}[htbp]
\centering
\begin{subfigure}{0.32\linewidth}
  \centering\includegraphics[width=\linewidth]{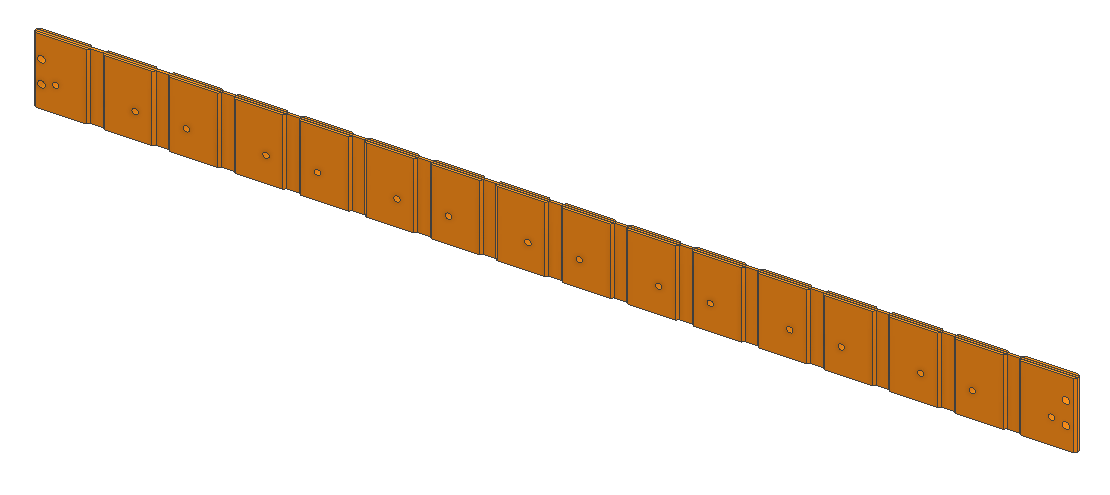}
  \caption{Flat pattern}\label{fig:soft-flat}
\end{subfigure}\hfill
\begin{subfigure}{0.30\linewidth}
  \centering\includegraphics[width=\linewidth]{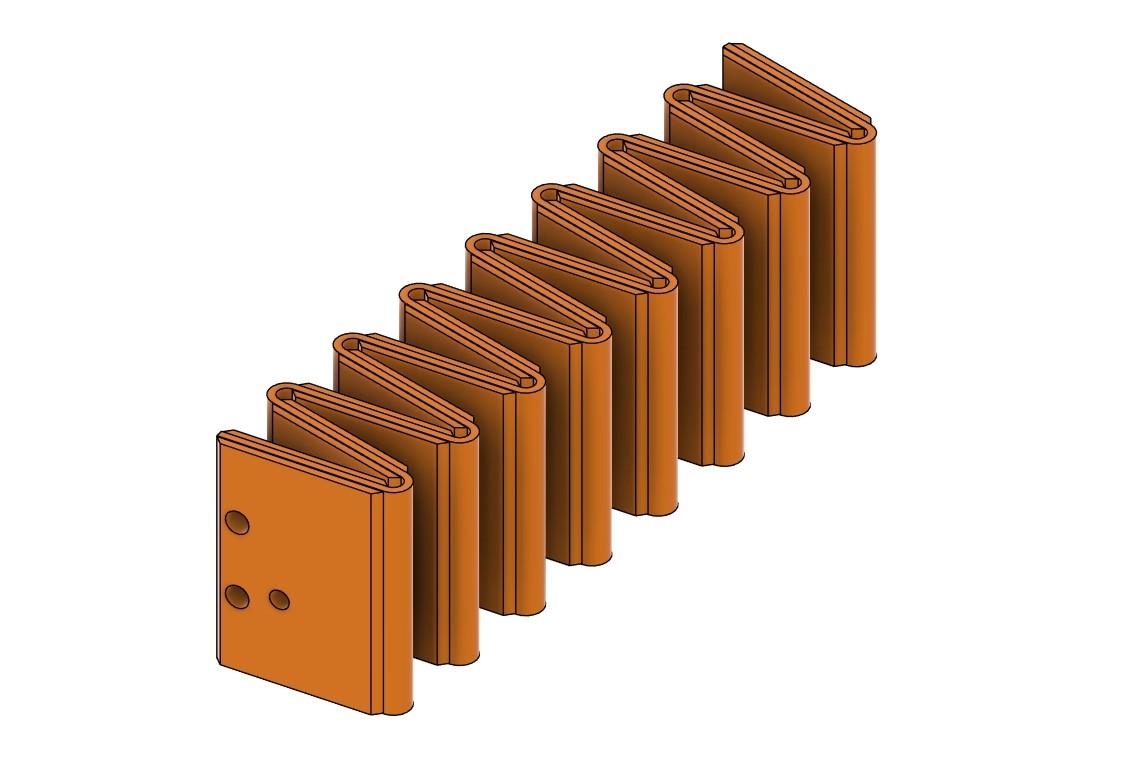}
  \caption{Folded}\label{fig:soft-folded}
\end{subfigure}\hfill
\begin{subfigure}{0.32\linewidth}
  \centering\includegraphics[width=\linewidth]{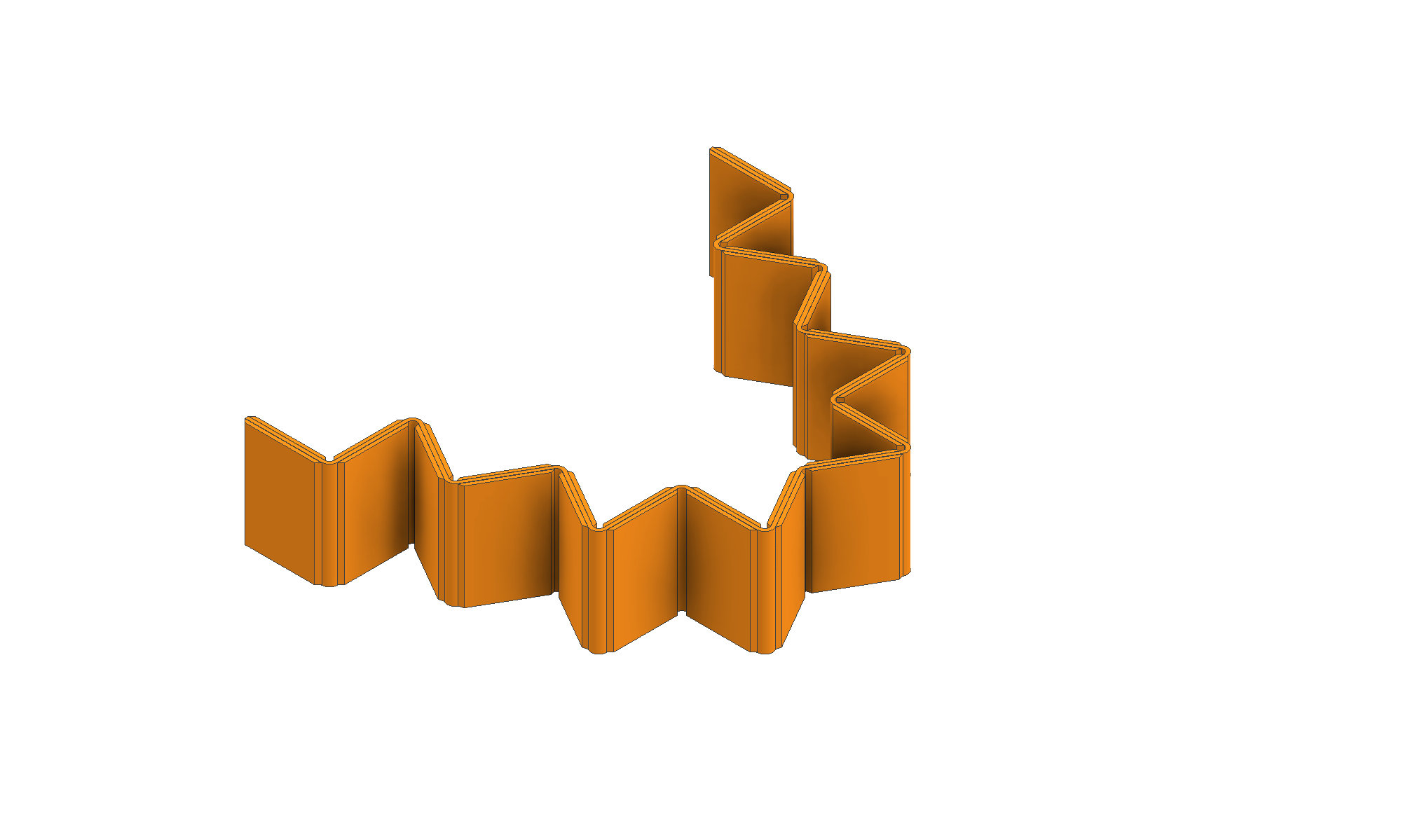}
  \caption{Deployed}\label{fig:soft-deployed}
\end{subfigure}
    \captionsetup{belowskip=-4mm}
\caption{The soft reconfigurable TPU structure. The flat-printed sheet
(a) folds into a compact state (b) for efficient flight and deploys into a
shock-absorbing envelope (c) when protection is required.}
\label{fig:soft}
\end{figure}

The soft bumper must fold compactly during unobstructed flight and deploy into a
shape that maximizes energy absorption in cluttered spaces. Two design choices
govern its behavior: \emph{material} and \emph{geometry}. Thermoplastic
polyurethane (TPU) is selected as a thermoplastic elastomer that combines
rubber-like elasticity with high elongation at break, tear strength, and impact
durability~\cite{TPUbasedComposites,TPUadditive}
(Table~\ref{tab:materials}). The geometry is an origami-inspired, flat-printed
pattern with pre-formed fold lines. Several folding patterns were prototyped and
evaluated for foldability and structural integrity; the final single-material
TPU pattern folds and unfolds repeatably without failure while exhibiting the
desired springiness.
Fig.~\ref{fig:soft} shows the flat pattern, the fully folded state, and the
deployed shock-absorbing state.

\begin{table}[htbp]
\caption{Typical mechanical properties of TPU (soft bumper) vs.\ PLA (rigid parts).}
\label{tab:materials}
\centering\renewcommand{\arraystretch}{1.15}
\footnotesize
\begin{tabular}{l c c c}
\toprule
\textbf{Property} & \textbf{Unit} & \textbf{TPU} & \textbf{PLA} \\
\midrule
Tensile strength      & MPa   & 20--60   & 50--70 \\
Elongation at break   & \%    & 300--700 & 2--10 \\
Tear strength         & kN/m  & 80--150  & 10--20 \\
Young's modulus       & MPa   & 10--1000 & 2000--3000 \\
Resilience (rebound)  & \%    & 30--60   & 5-15 \\
Impact strength       & --    & high     & low (brittle) \\
\bottomrule
\end{tabular}
\end{table}

\subsection{Rigid Shrouds and Tendon Actuation}
The rigid parts are 3D-printed in PLA, which retains its shape under structural
loads (Table~\ref{tab:materials}). A shroud is placed around each motor: it acts
as a beveled propeller guard that improves low-speed aerodynamics and shields the
blades, while also serving as the mounting interface for the soft bumper. A
bracket tangent to each shroud constrains the folded bumper into the desired
shape (Fig. 5); without such constraints the resilient TPU would spring into an arbitrary
form.

\begin{figure}[htbp]
\centering
\begin{subfigure}{0.485\linewidth}
  \centering\includegraphics[width=\linewidth]{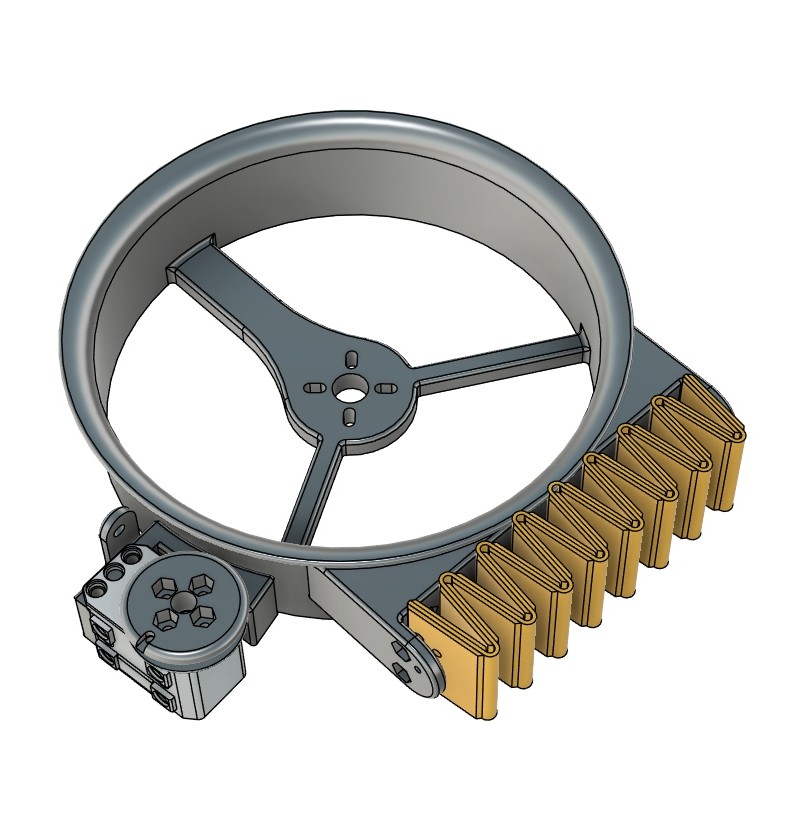}
  \caption{Folded (restrained)}\label{fig:act-folded}
\end{subfigure}\hfill
\begin{subfigure}{0.485\linewidth}
  \centering\includegraphics[width=\linewidth]{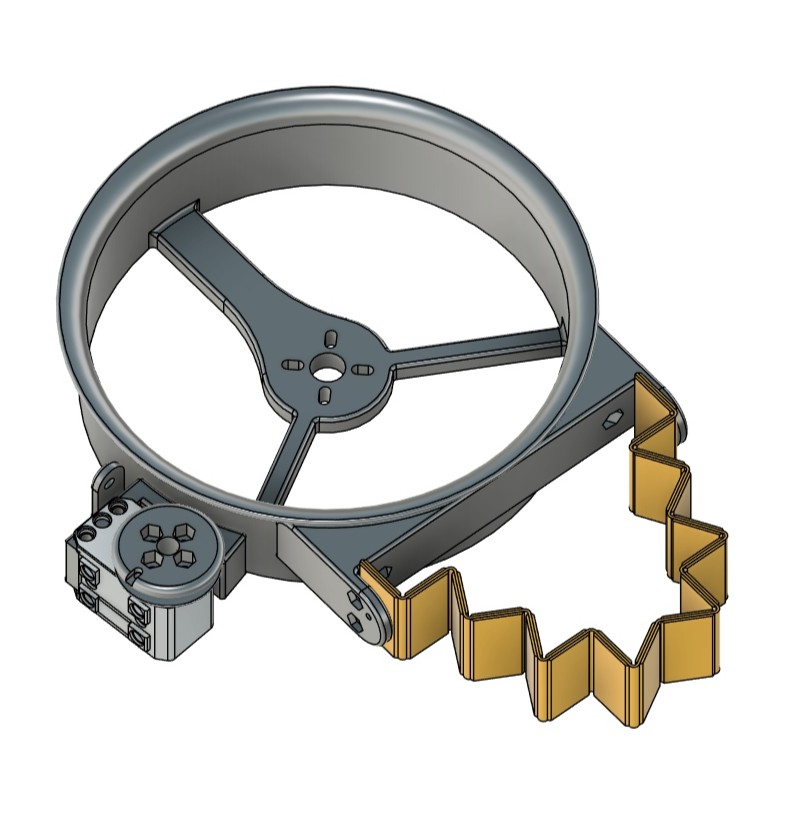}
  \caption{Deployed (released)}\label{fig:act-deployed}
\end{subfigure}
    \captionsetup{belowskip=-4mm}
\caption{The two states of a shroud--bumper assembly. A tendon wound on a
servo-driven pulley restrains the folded bumper (a); releasing the tendon lets
the elastic TPU self-deploy (b).}
\label{fig:actuation}
\end{figure}

Because the TPU naturally tends to spring open, actuation only needs to release
or restrain it. A string is routed through the folded bumper, wrapped around the
shroud, and wound onto a pulley on the servo shaft (Fig.~\ref{fig:actuation}).
Winding the string (counter-clockwise) pulls the bumper into its compact folded
state against the TPU's elastic restoring force; unwinding it (clockwise) loosens
the tendon and lets the bumper self-deploy. The four shrouds are identical and
independently actuated, one per arm.

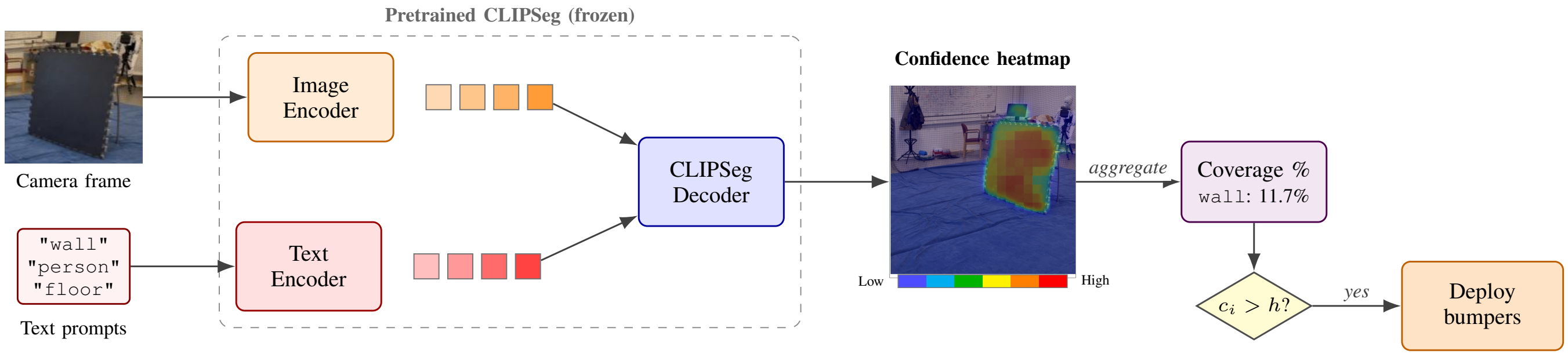
\begin{figure*}[t]
\centering
\resizebox{\textwidth}{!}{%
\begin{tikzpicture}[node distance=7mm and 12mm]
\node[img] (qimg) {\includegraphics[width=17mm]{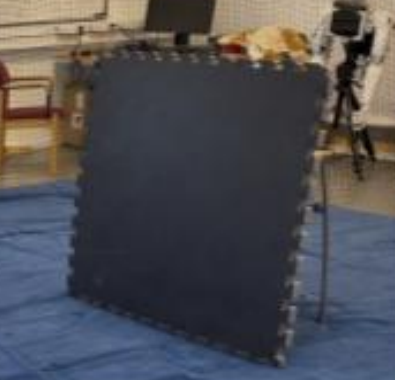}};
\node[below=0mm of qimg, font=\scriptsize] {Camera frame};
\node[prompt, below=8mm of qimg] (prompt)
     {\texttt{"wall"}\\\texttt{"person"}\\\texttt{"floor"}};
\node[above=-15mm of prompt, font=\scriptsize] {Text prompts};
\node[venc, right=13mm of qimg]   (venc) {Image\\Encoder};
\node[tenc, right=13mm of prompt] (tenc) {Text\\Encoder};

\foreach \i/\c in {0/30,1/45,2/60,3/78}
  \node[fsq, fill=orange!\c] (vf\i) at ($(venc.east)+(0.55+\i*0.42,0)$) {};
\foreach \i/\c in {0/25,1/40,2/58,3/74}
  \node[fsq, fill=red!\c] (tf\i) at ($(tenc.east)+(0.55+\i*0.42,0)$) {};

\node[dec2] at ($(vf3)!0.5!(tf3)+(2.2,0)$) (dec) {CLIPSeg\\Decoder};
\node[img, right=13mm of dec] (heat) {\includegraphics[width=23mm]{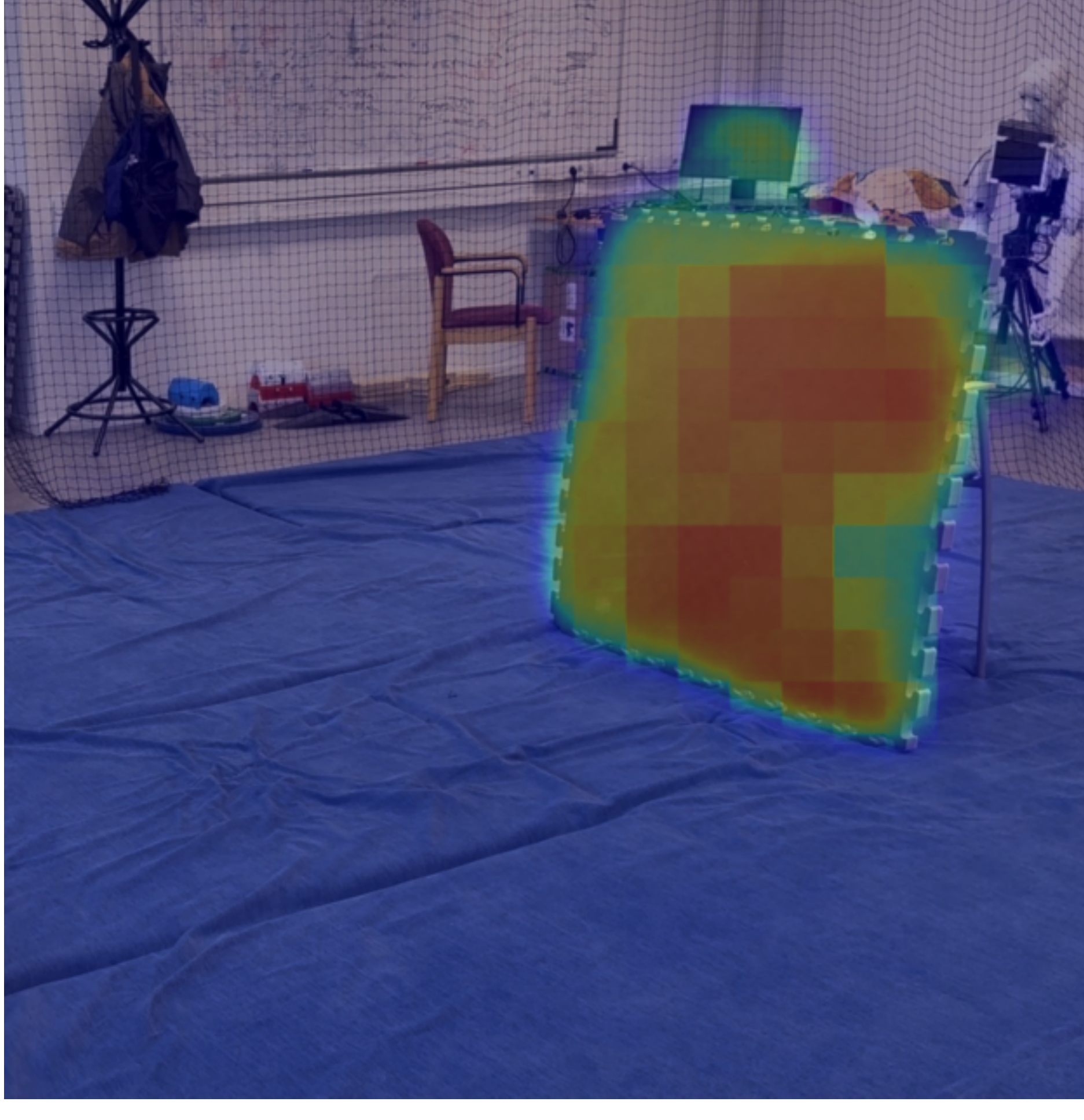}};
\node[above=0.8mm of heat, font=\scriptsize\bfseries] {Confidence heatmap};
\begin{scope}[shift={($(heat.south)+(0,-1.2mm)$)}]
 \foreach \i/\c in {0/{blue!70},1/{cyan},2/{green!70!black},3/{yellow},4/{orange},5/{red}}
   \fill[\c] (\i*3.5mm-10.5mm,0) rectangle ++(3.5mm,1.6mm);
 \draw[black!55,line width=0.3pt] (-10.5mm,0) rectangle (10.5mm,1.6mm);
 \node[font=\tiny, anchor=east] at (-11mm,0.8mm) {Low};
 \node[font=\tiny, anchor=west] at (11mm,0.8mm) {High};
\end{scope}

\node[cov, right=13mm of heat] (cov) {Coverage \%\\\scriptsize \texttt{wall}: 11.7\%};
\node[dec, below=6mm of cov] (thr) {$c_i>h$?};
\node[bmp, right=11mm of thr] (deploy) {Deploy\\bumpers};

\draw[arr] (qimg)   -- (venc);
\draw[arr] (prompt) -- (tenc);
\draw[arr] (vf3) -- (dec);
\draw[arr] (tf3) -- (dec);
\draw[arr] (dec)  -- (heat);
\draw[arr] (heat) -- node[lab,above]{aggregate} (cov);
\draw[arr] (cov)  -- (thr);
\draw[arr] (thr)  -- node[lab,above]{yes} (deploy);

\begin{scope}[on background layer]
\node[draw=black!45, dashed, rounded corners, inner sep=2mm,
      fit=(venc)(tenc)(vf3)(tf3)(dec),
      label={[glab,black!60]above:Pretrained CLIPSeg (frozen)}] {};
\end{scope}
\end{tikzpicture}}
    \captionsetup{belowskip=-4mm}
\caption{Language-driven perception pipeline. The onboard camera frame and the
free-text prompts (\texttt{wall}, \texttt{person}, \texttt{brick}) are embedded
by the frozen CLIP image and text encoders; the CLIPSeg decoder fuses the visual
and text features into a per-prompt confidence heatmap. Each heatmap is
aggregated into an image-coverage percentage that, once it exceeds the threshold
$h$ over the confirmation window, triggers deployment. Diagram style adapted from
a comparable open-vocabulary segmentation pipeline.}
\label{fig:clipseg}
\end{figure*}

\subsection{Finite Element Analysis of the Bumper}
\label{sec:fea}

The impact behavior of the bumper cannot be accurately described using simple
analytical models, as TPU exhibits near-incompressibility, hyperelasticity, and
strongly nonlinear deformation characteristics. Therefore, a set of assumptions
is introduced to enable a static nonlinear finite element analysis (FEA) using a
Mooney--Rivlin material model, providing an estimation of the deformation range,
peak reaction force, and structural integrity of the bumper
~\cite{FEforHyperelasticStructures,FEimpactHyperelastic,FiniteElementTechniques}.

To define a representative loading condition, a moderate impact scenario is
considered with a total UAV mass of \(m = 1.3\) kg, a pre-impact velocity of
\(v = 1.5\) m/s, and an estimated maximum bumper deformation of
\(d \approx 80\) mm. The corresponding kinetic energy is
\begin{equation}
\label{eq:impactEnergy}
E = \frac{1}{2}mv^2 \approx 1.6~\mathrm{J},
\end{equation}
which is converted into an equivalent average impact force approximation of
\begin{equation}
\label{eq:impactforce}
F = \frac{E}{d} \approx 20~\mathrm{N}.
\end{equation}

This load is assumed to be distributed between the two bumpers involved in the
contact event. The nonlinear finite element analysis is performed in the
Autodesk Fusion 360 simulation environment on a single deployed bumper. The TPU
material response is modeled using a two-parameter Mooney--Rivlin formulation,
where the strain energy density is expressed as
\begin{equation}
W_m = C_1(\bar I_1-3) + C_2(\bar I_2-3) + D_1(J-1)^2 ,
\end{equation}
where \(C_1\), \(C_2\), and \(D_1\) are experimentally determined material
constants. The bumper geometry is discretized using a tetrahedral mesh with
local refinement to 1 mm in the fold regions. Due to the deformation-dependent
stiffness of the hyperelastic material, the nonlinear equilibrium equations are
solved incrementally using Newton--Raphson iterations. The structure is then
loaded until the onset of failure to estimate the deformation capability,
reaction forces, and structural integrity under the applied impact-equivalent
loading.

\subsection{Perception and On-Demand Deployment}
\label{subsec: deployment}

\begin{figure}[h]
    \centering
    \includegraphics[width=1.0\linewidth]{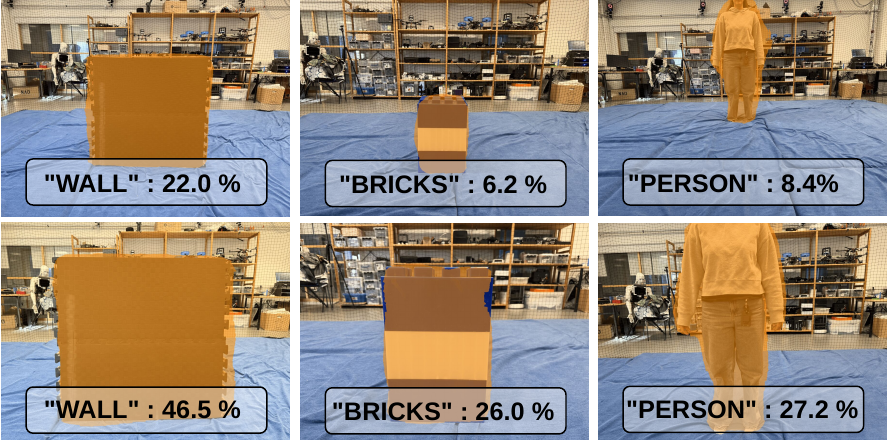}
        \captionsetup{belowskip=-4mm}
    \caption{Detection of different obstacles in the flight arena using the VLM. The figure shows the segmentation output and the corresponding percentage of image coverage for each detected object.}
    \label{fig:segmentations}
\end{figure}

To enable autonomous deployment, the CLIPSeg vision-language model assesses the environment surrounding the UAV during flight using a predefined set of textual prompts (e.g., ``wall,'' ``person''). Frames captured by a Raspberry Pi camera connected to a Raspberry Pi Zero~2~W board are streamed over Wi-Fi to a FastAPI service hosted on a GPU workstation (RTX~4080) running CLIPSeg. For each prompt, the model generates a segmentation map indicating the image regions most strongly associated with the concept, visualized as a heatmap where higher-probability regions appear in red. This mask is aggregated into an image coverage percentage, a compact proxy for obstacle proximity, and returned as a JSON payload (e.g.\ \texttt{\{"wall": 11.7\}}), as illustrated in Fig.~\ref{fig:clipseg}. In the experiments reported here, the obstacle class is restricted to a black foam wall; however, the underlying model is prompt-agnostic and can, in principle, segment any object category specified at inference time via free-text input, without retraining. To demonstrate this generality, Fig.~\ref{fig:segmentations} presents segmentation outputs for three representative obstacle types: a wall, a brick structure, and a person, chosen to reflect typical environmental complexity in the vehicle's operating environment. For each frame, the VLM returns a coverage percentage corresponding to the queried prompt, providing a lightweight, interpretable signal of obstacle presence that can be incorporated into the UAV's decision-making pipeline.

The onboard controller continuously evaluates these coverage values against predefined thresholds, maintaining a short detection history to reduce noise sensitivity. The exoskeleton is deployed when any monitored coverage exceeds a high threshold~$h$ over a short confirmation window~$w_s$ (two consecutive frames), and retracted once all coverages fall below a low threshold~$\ell$, with a cooldown period~$T_c$ preventing command chatter. Commands are sent over a serial connection to the flight controller, closing the perception-control loop; new behaviors can be added simply by editing the text prompts. The full pipeline and system architecture are shown in Figs.~\ref{fig:clipseg} and~\ref{fig:arch}, respectively. End-to-end latency is dominated by Wi-Fi transmission ($\sim$80~ms) and image resizing ($\sim$40~ms), with CLIPSeg inference contributing only $\sim$10~ms, for a typical per-frame budget of 60--150~ms.

\begin{figure}[h]
\centering
\begin{subfigure}{0.9\linewidth}
  \centering\includegraphics[width=\linewidth]{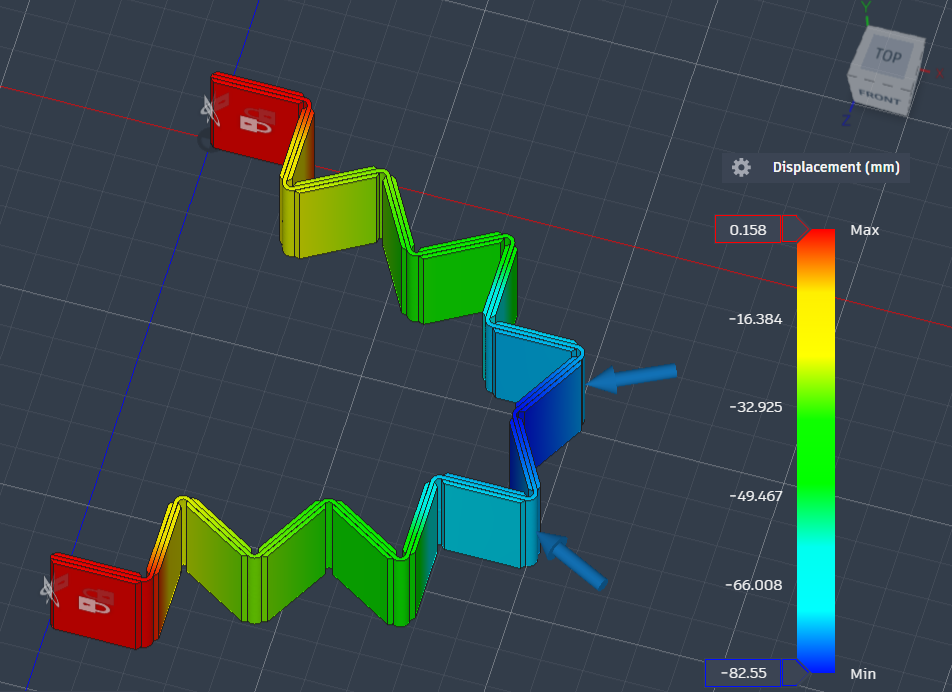}
  \caption{Displacement ($\approx82.6$\,mm)}\label{fig:fea-disp}
\end{subfigure}\\[2pt]
\begin{subfigure}{0.9\linewidth}
  \centering\includegraphics[width=\linewidth]{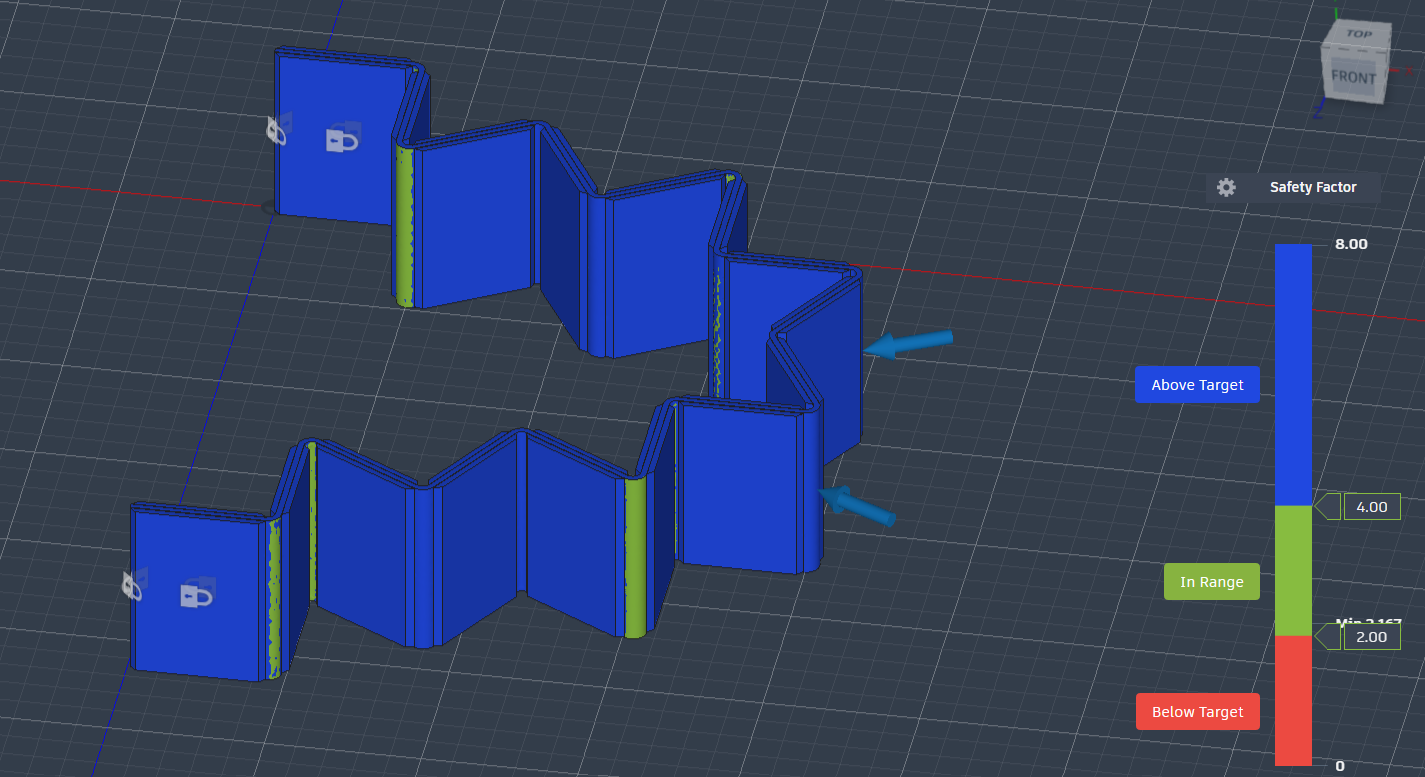}
  \caption{Safety factor ($\geq 4$)}\label{fig:fea-sf}
\end{subfigure}

\captionsetup{belowskip=-4mm}
\caption{Static hyperelastic FEA of one deployed bumper at the $10$~N load case.
Maximum travel is reached (a) while the safety factor stays above the target of
$4$ throughout the structure (b).}
\label{fig:fea}
\end{figure}

\section{EXPERIMENTAL VALIDATION}

\subsection{FEA Results}
The bumper was loaded incrementally until reaching its physical travel limit.
Under two perpendicular $3.5$~N loads 
%
%
($7$~N total) it deflects $\approx56$~mm
with equivalent strain concentrated at the fold lines, the expected failure
sites. Increasing to two $5$~N loads ($10$~N total) reaches the maximum physical
travel of $\approx82.6$~mm at an equivalent strain of $\sim0.5$
(Fig.~\ref{fig:fea}). The safety factor across the structure remains above the
target value of $4$, indicating structural integrity even at maximum
deformation. The static analysis thus suggests a per-bumper capacity of
$\approx10$~N ($\approx20$~N for a two-bumper contact), consistent with the
estimate in~\eqref{eq:impactforce}. However, as shown below in (Fig. \ref{fig:impactForce1}), the real, dynamic structure
sustains substantially higher transient forces than this conservative static
bound.

\subsection{Experimental Setup}

\begin{figure}[htbp]
\centering
\includegraphics[width=\linewidth]{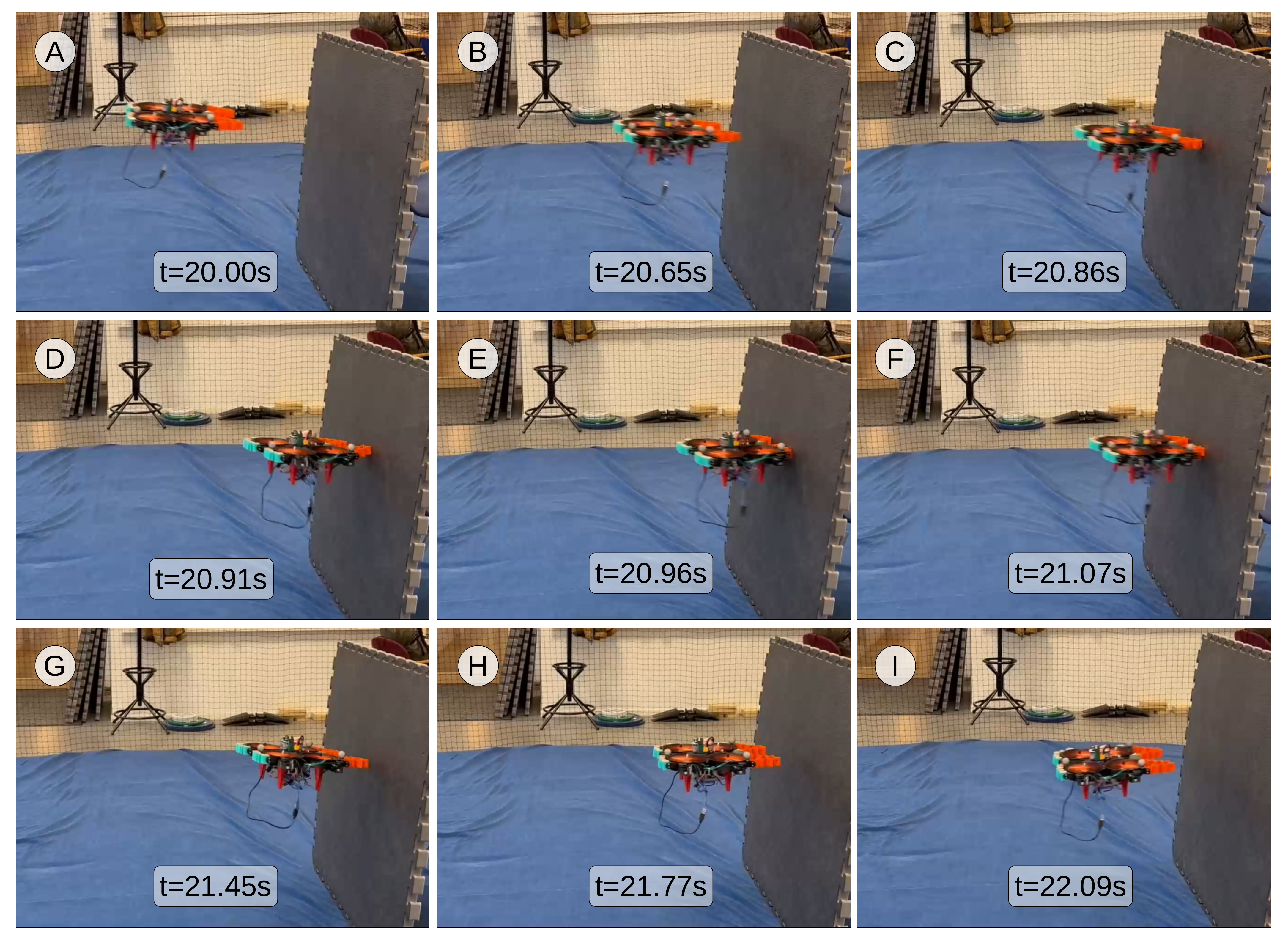}
    \captionsetup{belowskip=-4mm}
\caption{Side-view snapshots of a collision with the deployed bumper. Contact
occurs between frames D and G ($\sim$540\,ms); the UAV rebounds and resumes
flight.}
\label{fig:snapshots}
\end{figure}

Collision tests were performed in a lab environment
, tracked by an 18-camera Vicon T40S motion-capture
system providing position, velocity, and acceleration ground truth. A vertical
wall of foam mat blocks served as a repeatable obstacle. The pilot flew the UAV
manually into the wall at moderate speed. Each experiment was run twice under
otherwise identical conditions: once with the exoskeleton \emph{folded} (inactive)
and once with it autonomously \emph{deployed}. From the tracked motion we compute
the peak impact force $F_{\max}=m\,a_{\max}$ (with $m=1.3$~kg), the peak
acceleration, the pre-impact velocity (used to match comparable impacts), the
impact duration (window where $|a|$ exceeds $10\%$ of its peak). Snapshots of the flight with the deployed exoskeleton can be found in Fig.~\ref{fig:snapshots}.

\begin{figure}
\centering
\includegraphics[width=0.9\linewidth]{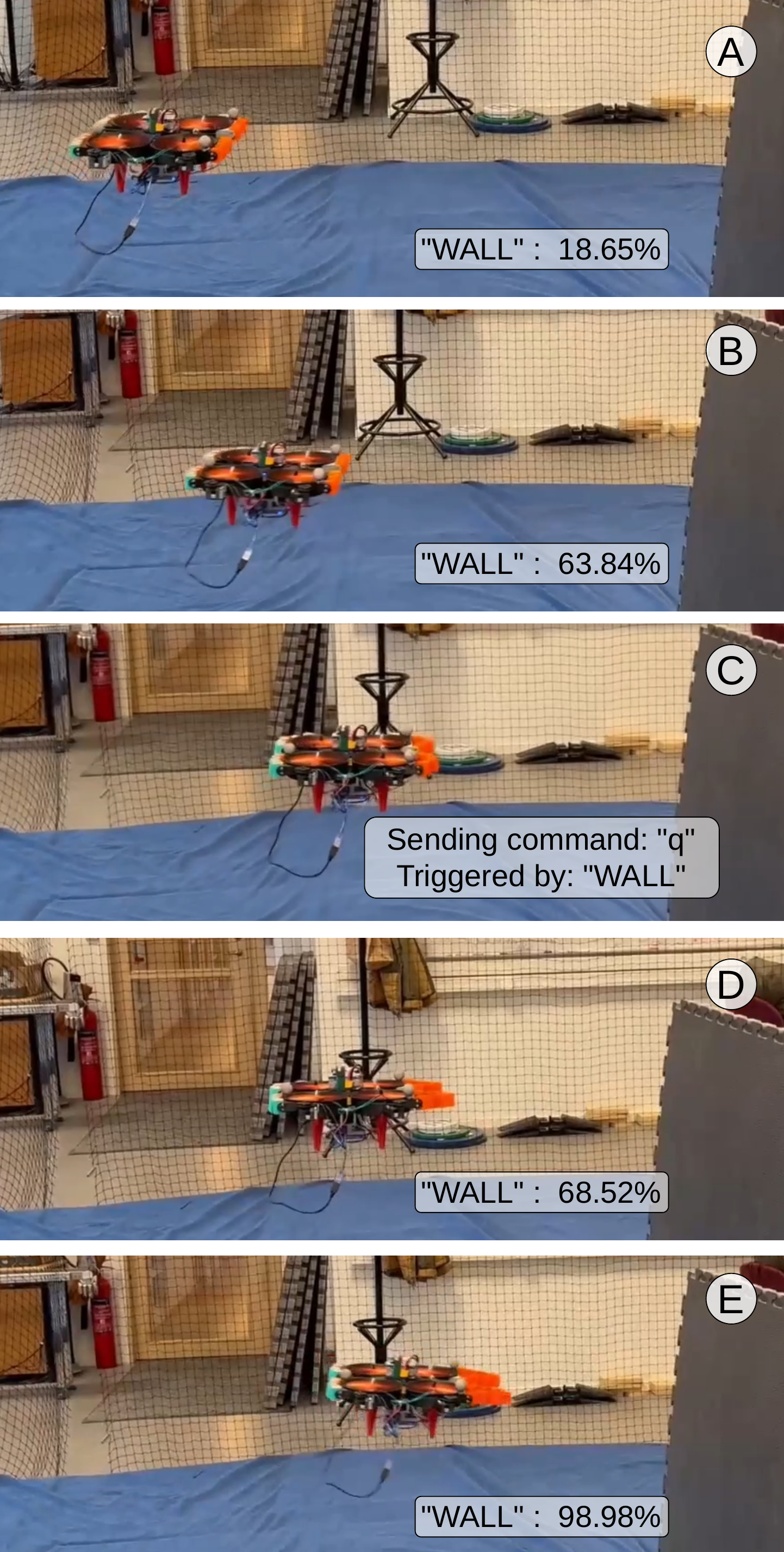}
    \captionsetup{belowskip=-4mm}
\caption{Autonomous deployment sequence with CLIPSeg coverage values for the
\texttt{wall} prompt. Deployment is triggered once the value exceeds the
threshold over the confirmation window (snapshot C).}
\label{fig:deploy}
\end{figure}

\begin{figure*}[t]
\centering
\begin{subfigure}{0.485\linewidth}
  \centering\includegraphics[width=\linewidth]{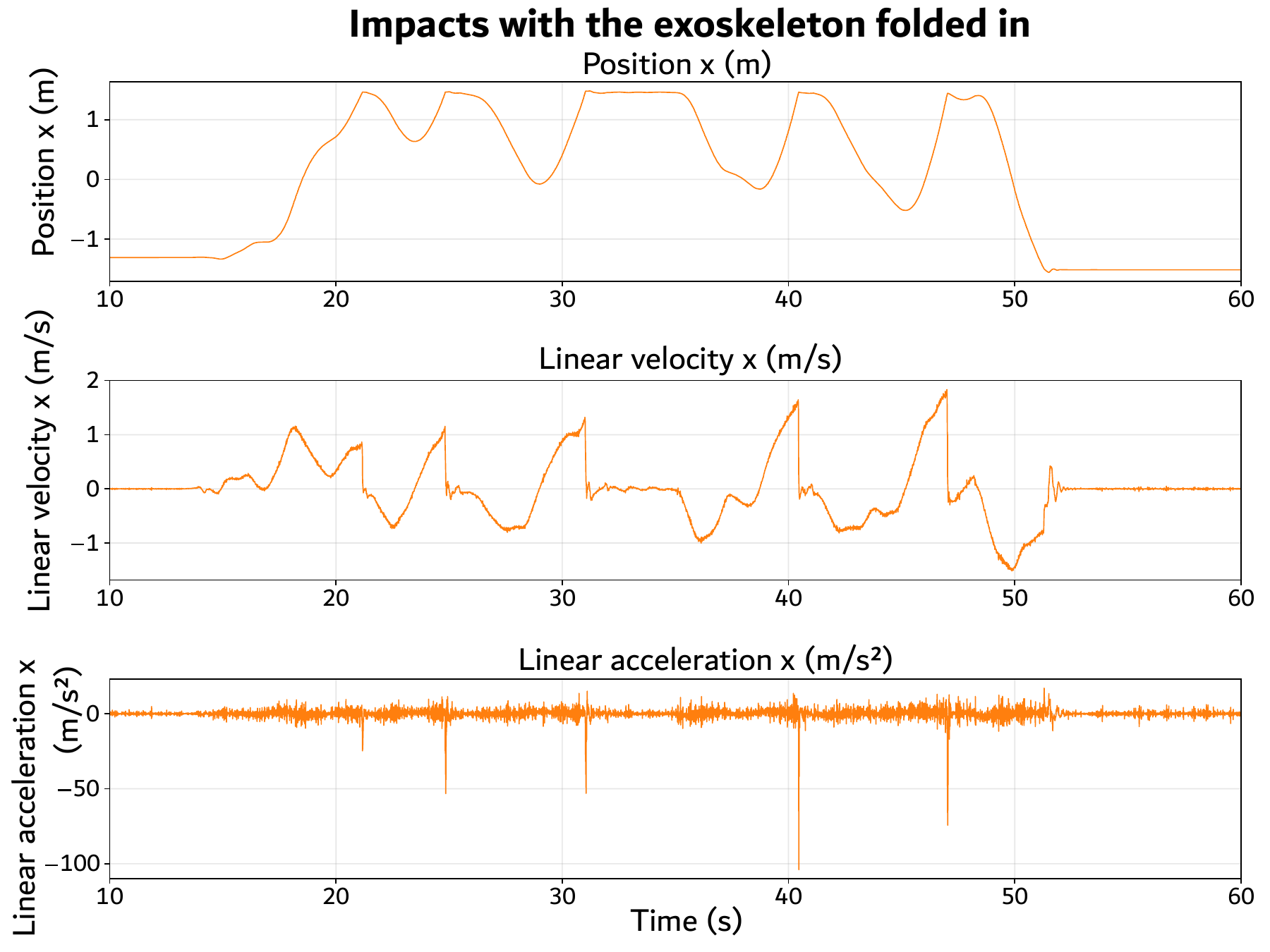}
  \caption{Exoskeleton folded (inactive)}\label{fig:flight-folded1}
\end{subfigure}\hfill
\begin{subfigure}{0.485\linewidth}
  \centering\includegraphics[width=\linewidth]{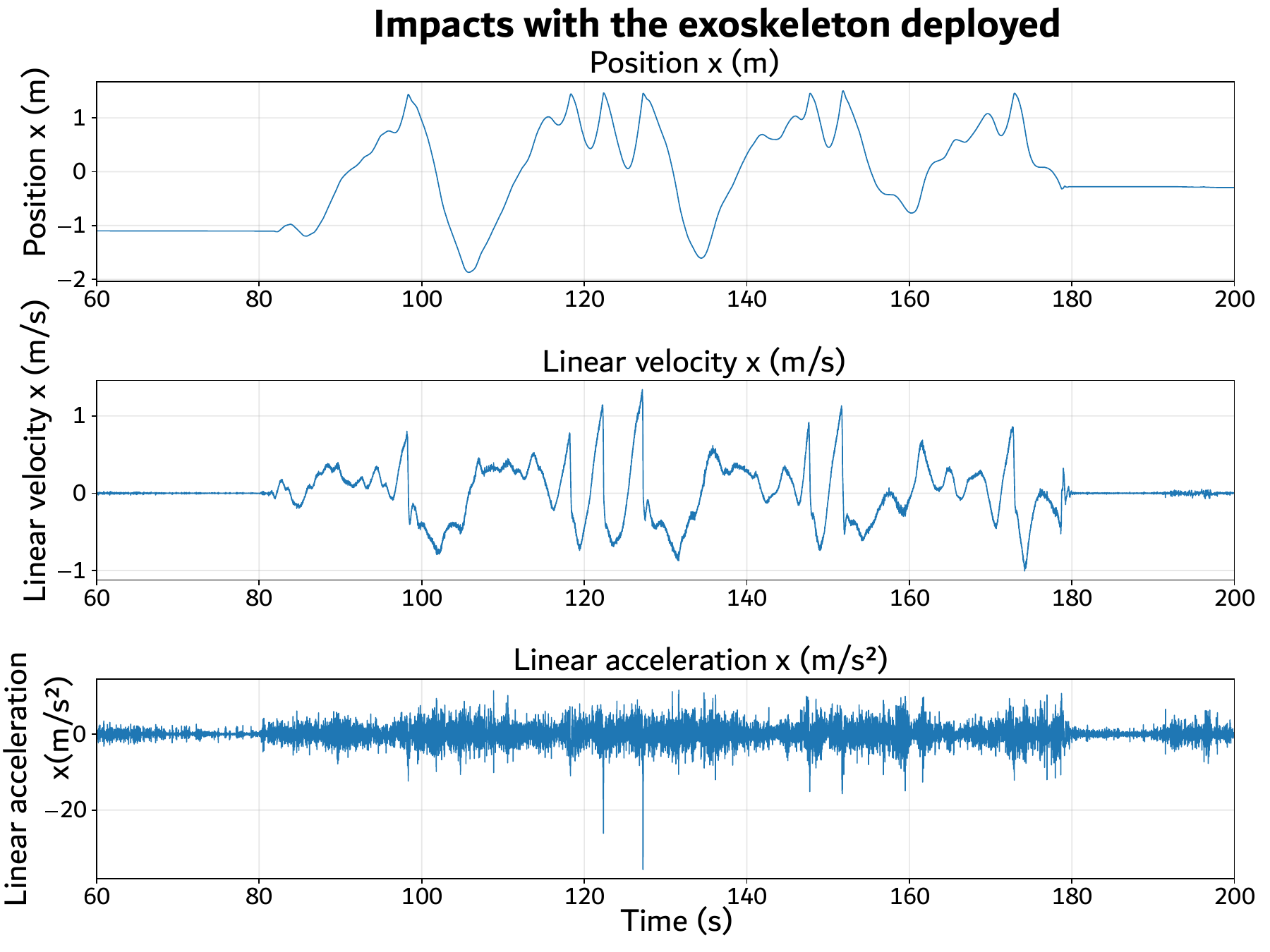}
  \caption{Exoskeleton deployed}\label{fig:flight-deployed1}
\end{subfigure}
\caption{Position, velocity, and acceleration over a full test flight with the
exoskeleton folded (a) and deployed (b). Each position peak corresponds to a wall
impact; the deployed case shows sharper rebounds and smoother velocity
transitions.}
\label{fig:flights}
\end{figure*}

\subsection{Experimental Results}

Fig.~\ref{fig:deploy} shows snapshots of an autonomous deployment together with
the CLIPSeg risk values for the \texttt{wall} prompt. Far from the obstacle the
coverage is low ($18.65$\%); as the UAV approaches, it rises until it exceeds the
high threshold $h=40\%$ over two consecutive frames, triggering deployment. After
the vehicle retreats and the coverage drops below $\ell=30\%$, the bumpers fold
back automatically, ready for the next event.

\begin{figure*}[t]
\centering
\begin{subfigure}{0.52\linewidth}
  \centering\includegraphics[width=\linewidth]{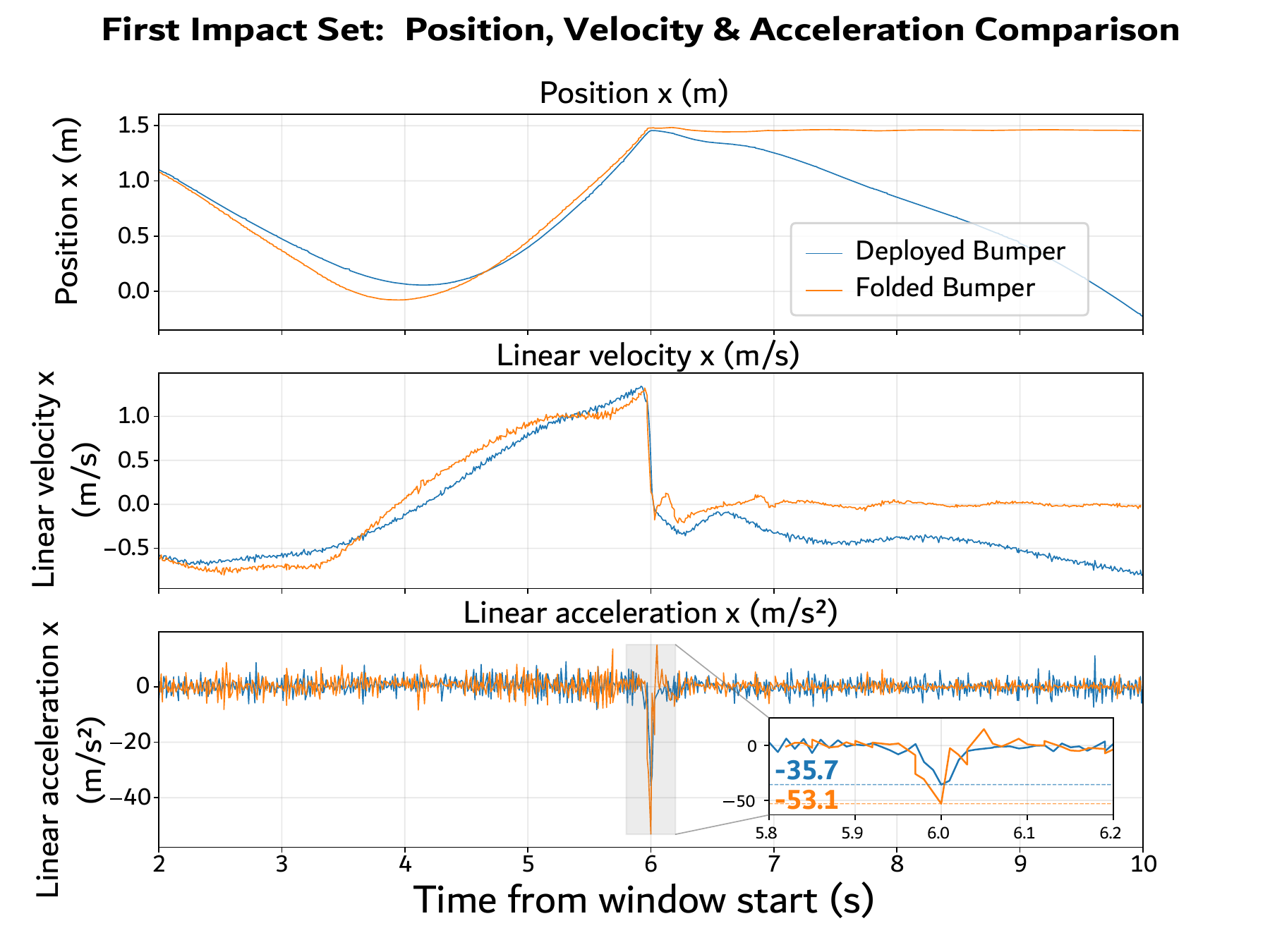}
  \caption{First impact set: similar velocities before impact. }\label{fig:flight-folded2}
\end{subfigure}\hfill
\begin{subfigure}{0.47\linewidth}
  \centering\includegraphics[width=\linewidth]{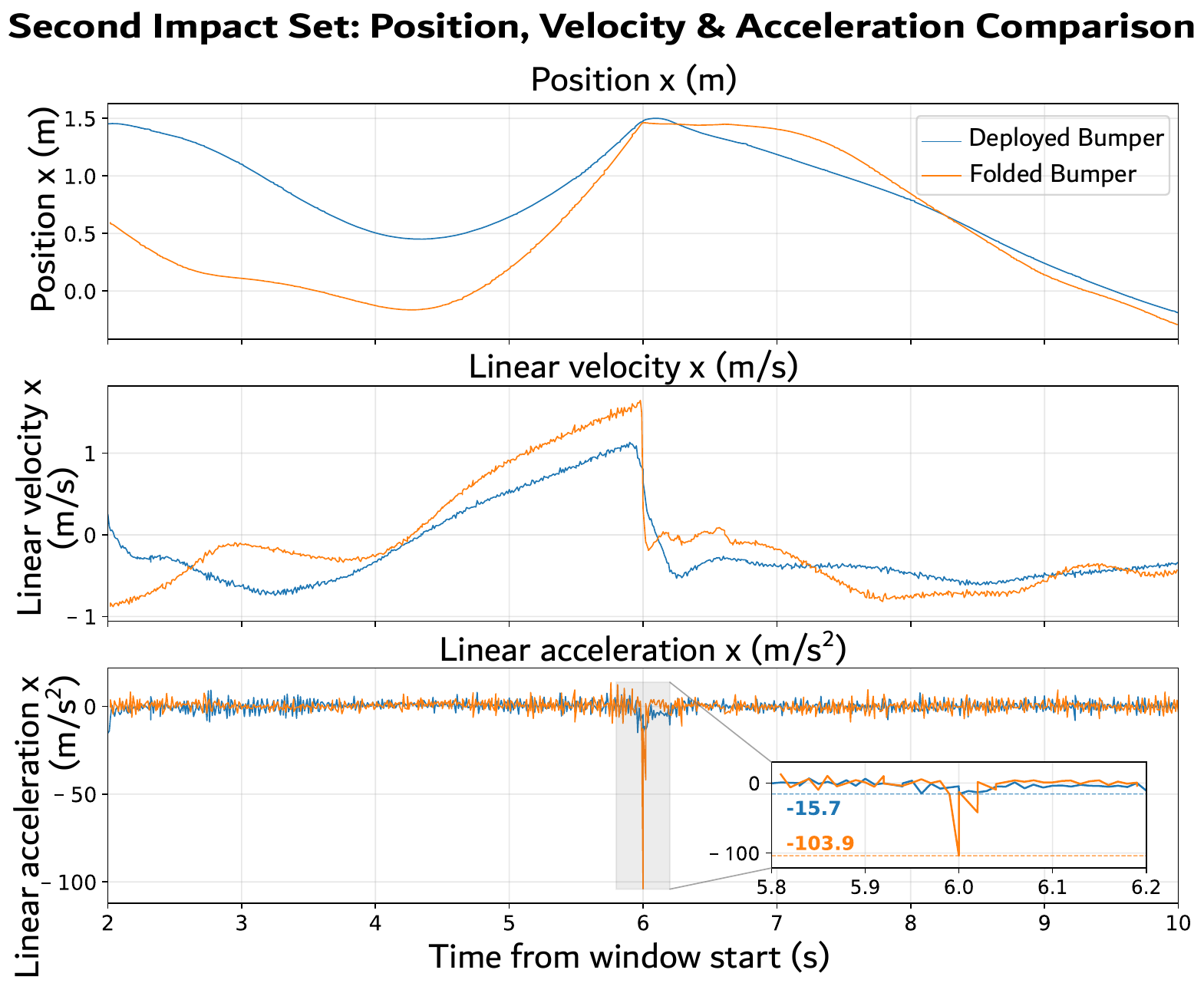}
  \caption{Second impact set: slightly different velocities before impact. }\label{fig:flight-deployed2}
\end{subfigure}
\caption{Velocity-matched comparison for two different impact sets: position, velocity, and
acceleration for folded vs.\ deployed exoskeleton. The inset highlights the reduced
peak acceleration with the bumper deployed in both cases.}
\label{fig: compare}
\end{figure*}

\begin{figure*}[t]
\centering
\begin{subfigure}{0.49\linewidth}
  \centering\includegraphics[width=\linewidth]{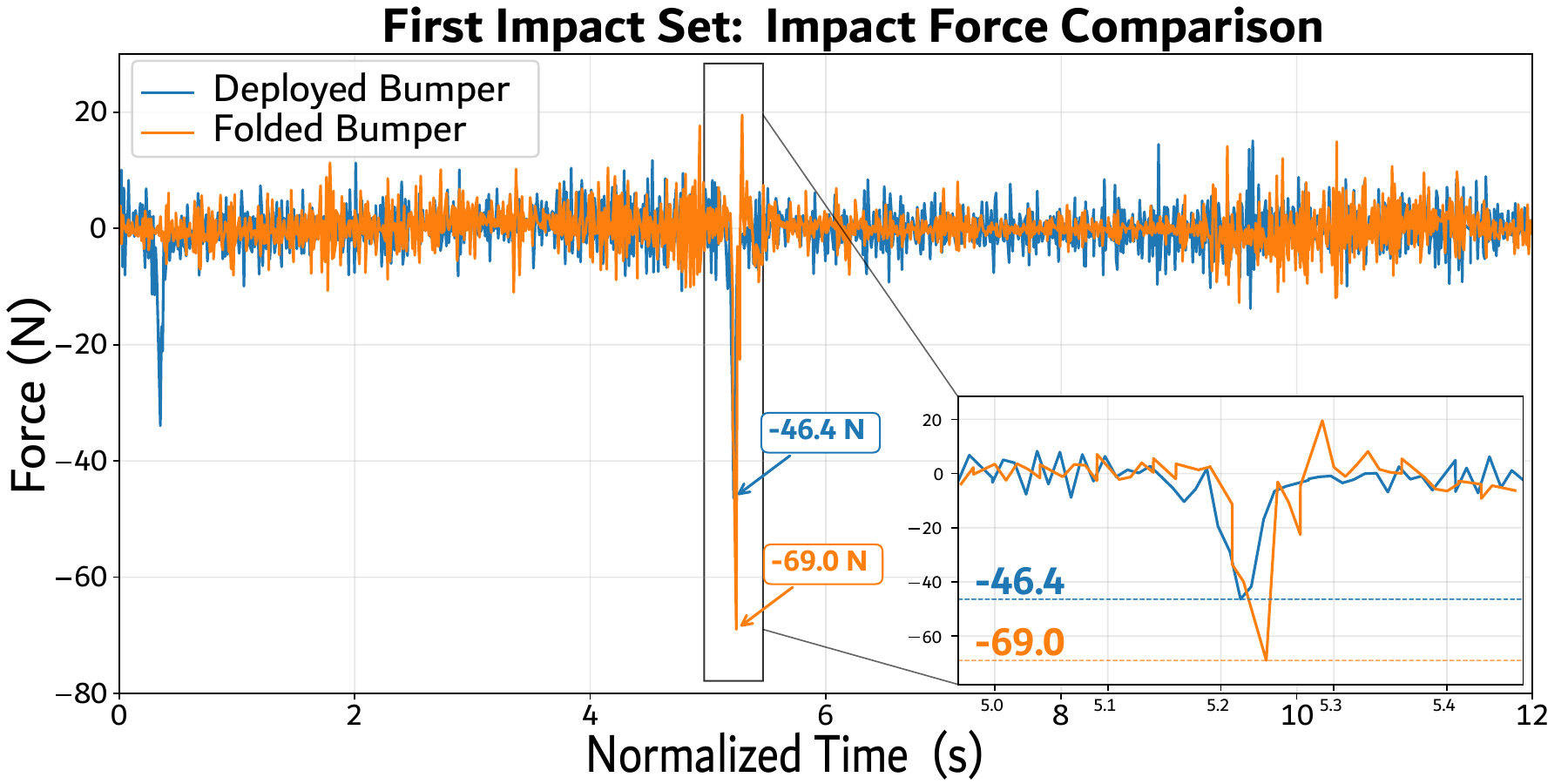}
  \caption{First impact set: similar velocities before impact. }\label{fig:impactForce1}
\end{subfigure}\hfill
\begin{subfigure}{0.49\linewidth}
  \centering\includegraphics[width=\linewidth]{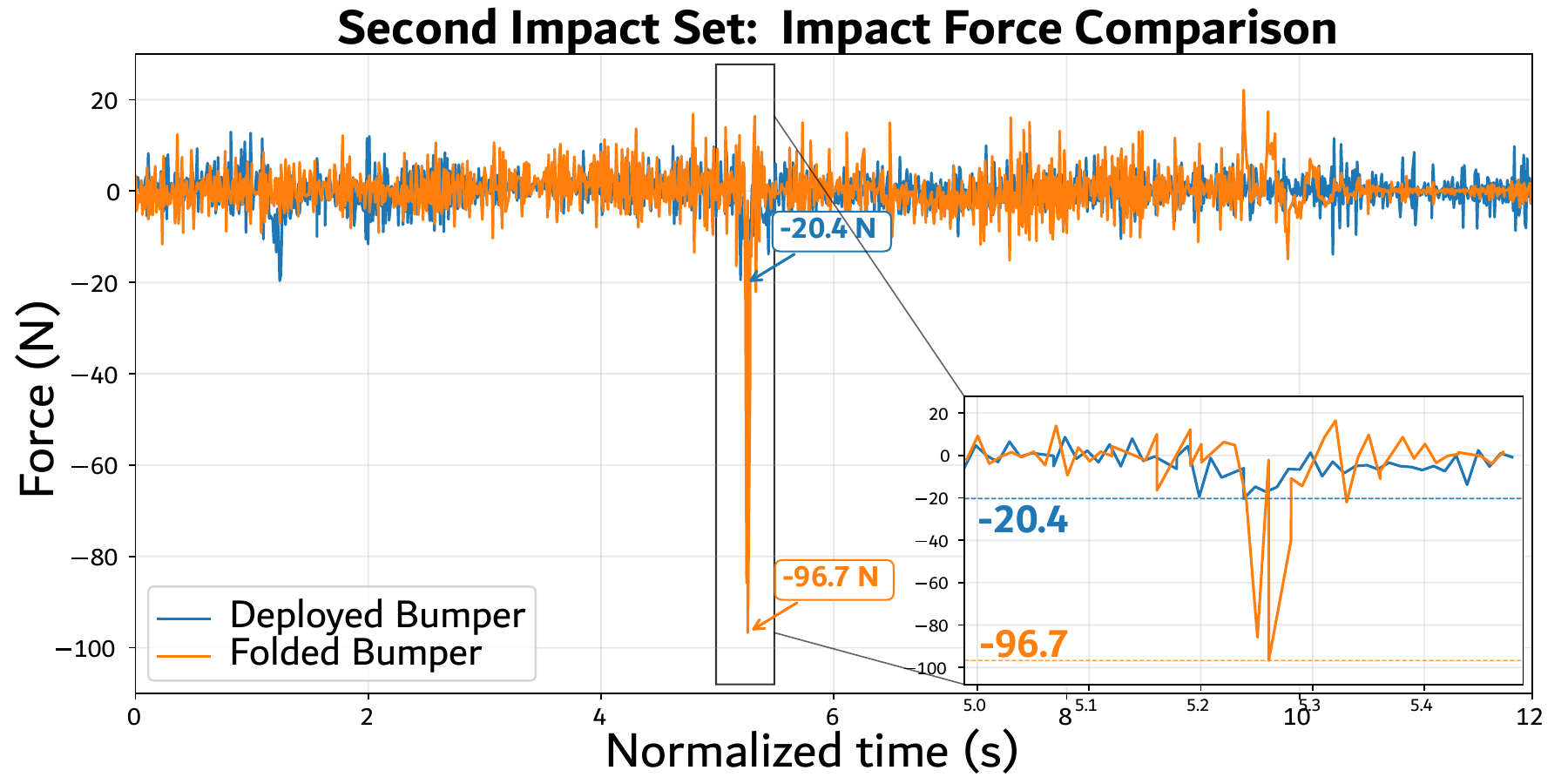}
  \caption{Second impact set: slightly different velocities before impact. }\label{fig:impactForce2}
\end{subfigure}
\caption{The deployed exoskeleton lowers the peak reaction force in both cases. The difference in the reaction force is much more visible at the second impact where the velocities differ.}
\label{fig: impactForce}
\end{figure*}

\begin{figure*}[t]
\centering
\begin{subfigure}{0.49\linewidth}
  \centering\includegraphics[width=\linewidth]{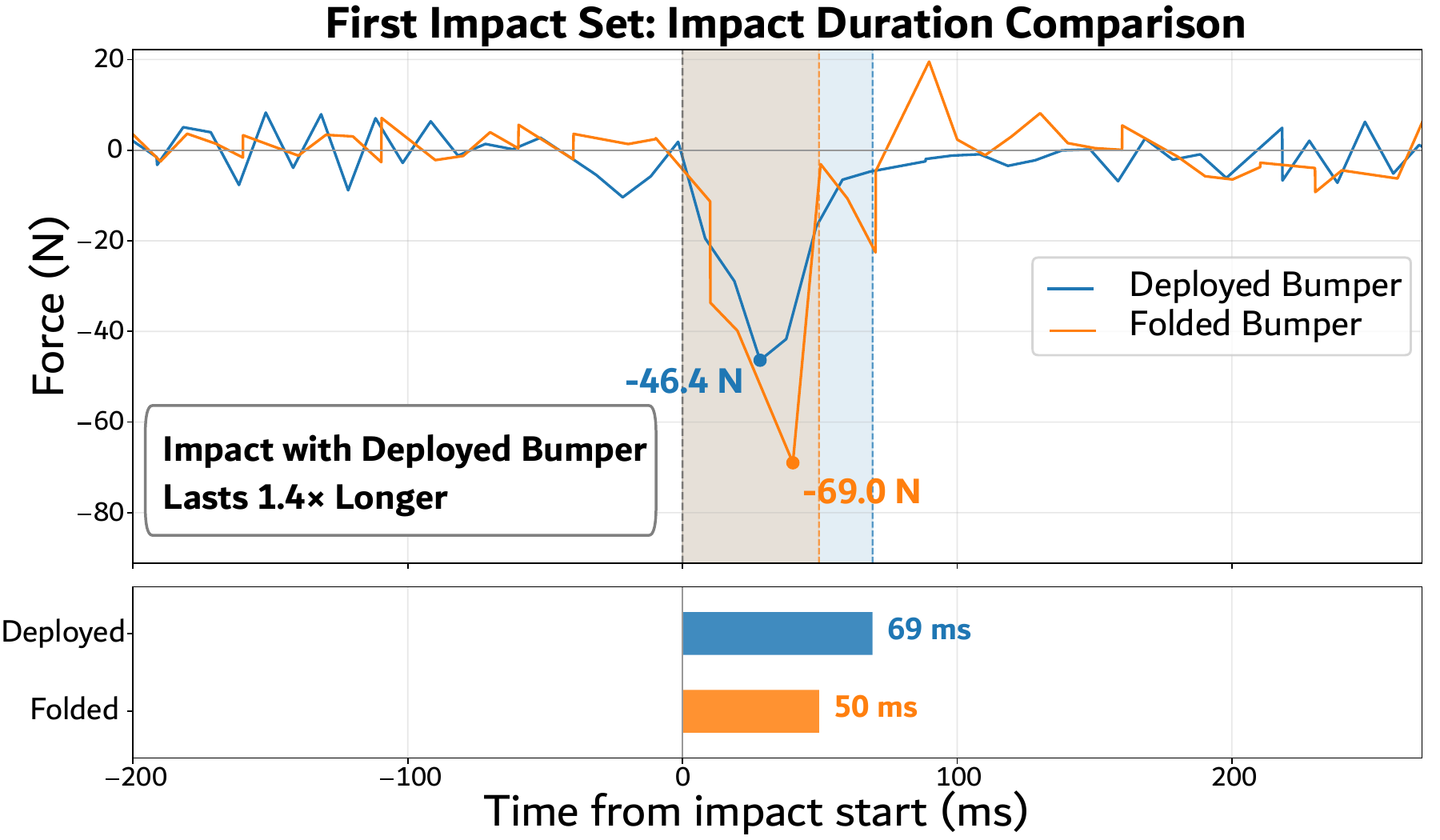}
  \caption{First impact set: similar velocities before impact. }\label{fig:impactDuration1}
\end{subfigure}\hfill
\begin{subfigure}{0.49\linewidth}
  \centering\includegraphics[width=\linewidth]{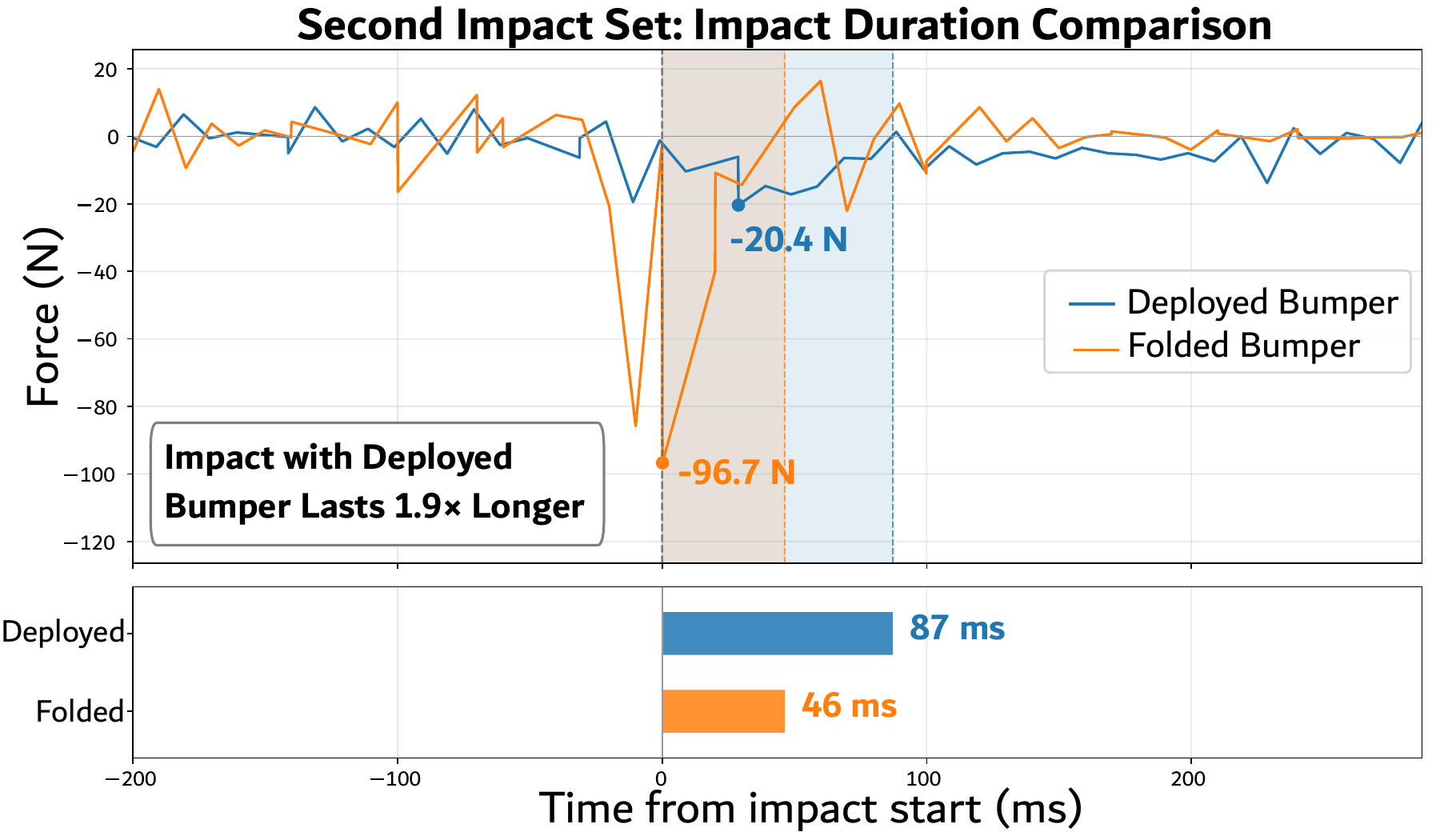}
  \caption{Second impact set: slightly different velocities before impact. }\label{fig:impactDuration2}
\end{subfigure}
    \captionsetup{belowskip=-4mm}
\caption{The deployed exoskeleton broadens the contact window, evidencing better energy dissipation.}
\label{fig: impactDuration}
\end{figure*}


Fig.~\ref{fig:flights} presents the complete flight trajectories recorded for both configurations (i.e., folded and deployed), where each local peak in the position profile corresponds to a discrete wall-impact event. When the exoskeleton is deployed, these position peaks are followed by a clear rebound, while the corresponding velocity transitions exhibit a smoother, more gradual profile. This behavior is consistent with a spring-damper interaction, in which the compliant structure engages the obstacle progressively through three distinct phases: an initial compression phase as the bumper deforms and absorbs kinetic energy, a brief full-contact phase at maximum deformation, and a subsequent decompression phase during which stored elastic energy is released and the vehicle rebounds away from the wall. In contrast, with the bumper folded, each velocity peak drops almost instantaneously to zero at the moment of contact, indicating an abrupt loss of momentum characteristic of a near-rigid, largely inelastic collision. This qualitative contrast between the two configurations provides further evidence, beyond the aggregate force and acceleration metrics, that the deployed exoskeleton meaningfully alters the collision dynamics rather than simply attenuating peak values.

This rebound behavior is further corroborated visually in Fig.~\ref{fig:snapshots}, which presents a side-view frame sequence of a representative collision with the exoskeleton deployed. The sequence captures the complete contact event, from initial engagement with the wall through compression, momentary full contact, and subsequent decompression, providing a direct visual counterpart to the spring--damper dynamics inferred from the position and velocity profiles above. The contact phase lasts approximately $\sim$540\,ms (frames D--G), after which the UAV cleanly separates from the wall and continues its flight along a stable trajectory. This combination of a finite, compliant contact duration and a clean post-impact recovery constitutes the expected and desired behavior of a protective exoskeleton, confirming that the structure not only attenuates peak loads but also enables the vehicle to disengage from the obstacle without loss of controllability.

To enable a fair comparison between configurations, impacts are aligned by matching pre-impact speed rather than by trial order alone, as shown in Fig.~\ref{fig: compare}. For the first velocity-matched set, corresponding to a pre-impact speed of approximately $1.3$~m/s, the deployed exoskeleton yields a marked reduction in peak acceleration relative to the folded configuration and allows the UAV to recover to its pre-impact position more rapidly (Fig.~\ref{fig:snapshots}). This faster recovery, combined with the reduced peak acceleration, indicates that a substantial fraction of the impact energy is absorbed and subsequently returned to the vehicle through elastic rebound, rather than being transmitted directly into the airframe.

\/*\begin{figure}[h!]
\centering
\includegraphics[width=0.99\linewidth]{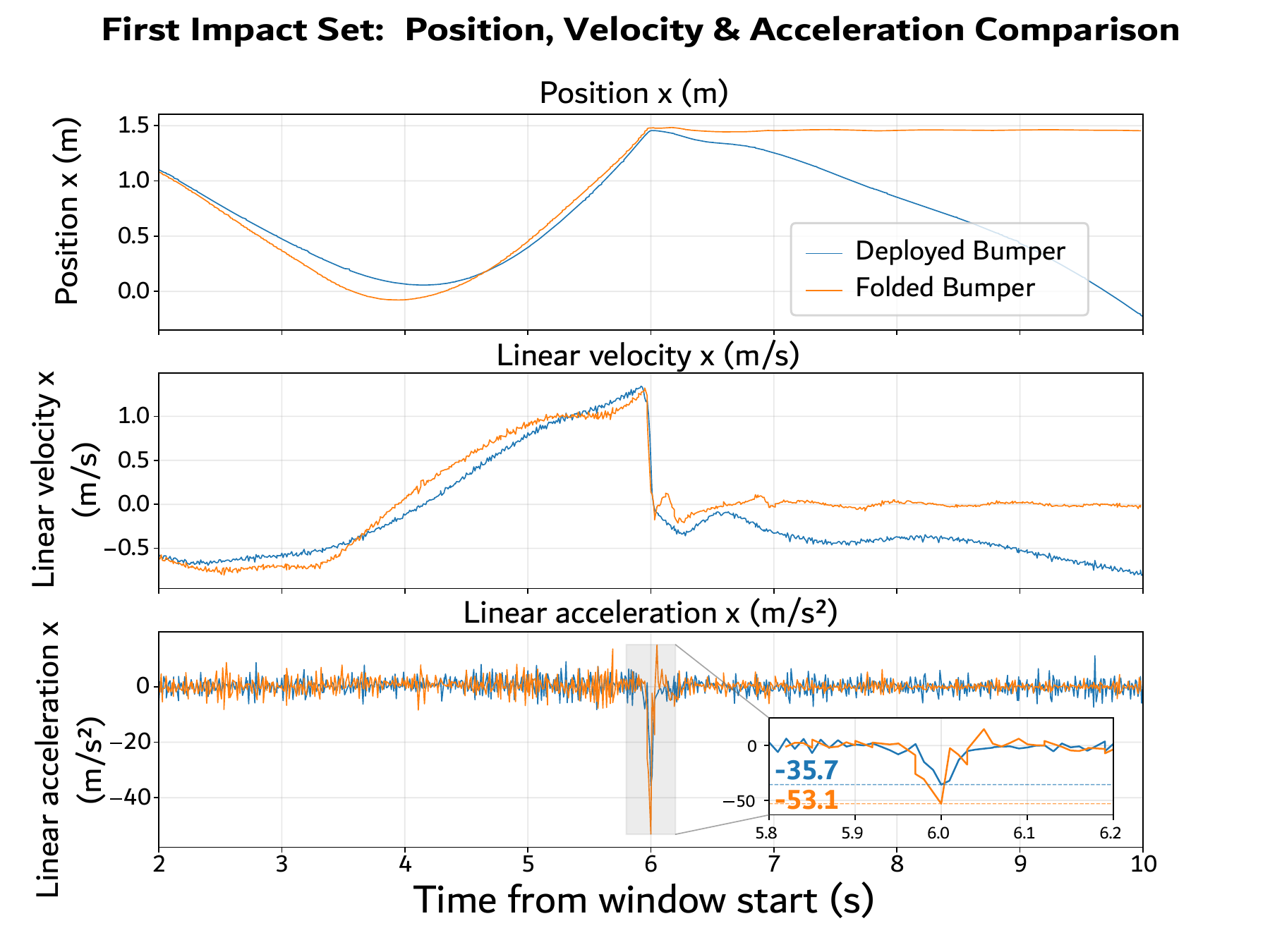}
\caption{Velocity-matched comparison (first impact set): position, velocity, and
acceleration for folded vs.\ deployed exosuit. The inset highlights the reduced
peak acceleration with the bumper deployed.}
\label{fig:compare}
\end{figure}
*/
This interpretation is further supported by the corresponding reaction force measurements (Fig.~\ref{fig: impactForce}), which show a reduction in peak reaction force from $69.0$~N in the folded configuration to $46.4$~N with the exoskeleton deployed, corresponding to a $33\%$ reduction. In addition to this reduction in magnitude, the force pulse in the deployed configuration is visibly broader in duration, indicating a longer and more gradual contact interaction consistent with the compliant, energy-absorbing behavior of the exoskeleton rather than the abrupt, near-impulsive contact observed in the folded case.

\/*\begin{figure}[h!]
\centering
\begin{subfigure}{0.99\linewidth}
  \centering\includegraphics[width=\linewidth]{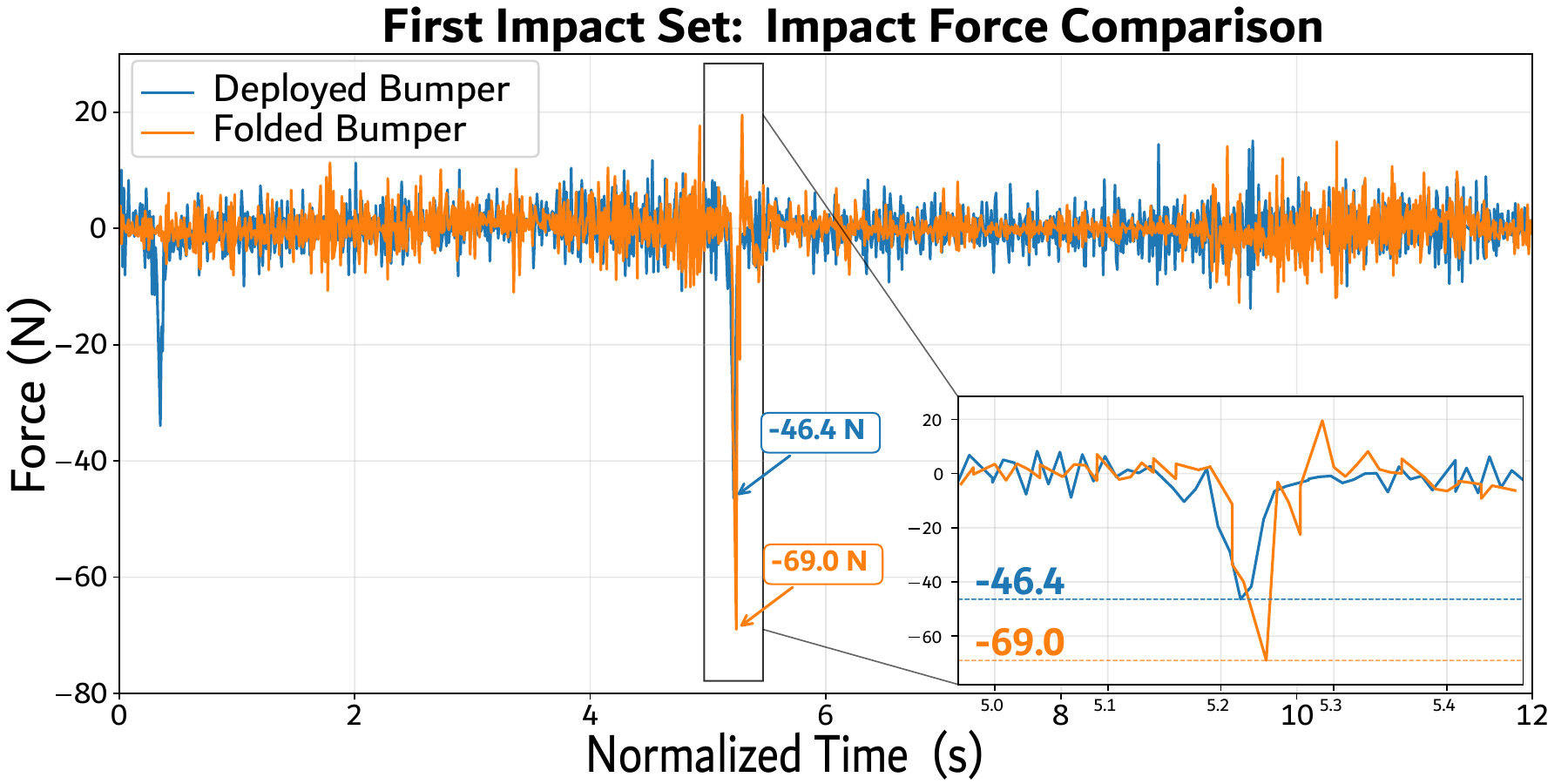}
  \caption{Impact force}\label{fig:force-a}
\end{subfigure}\\[2pt]
\begin{subfigure}{0.99\linewidth}
  \centering\includegraphics[width=\linewidth]{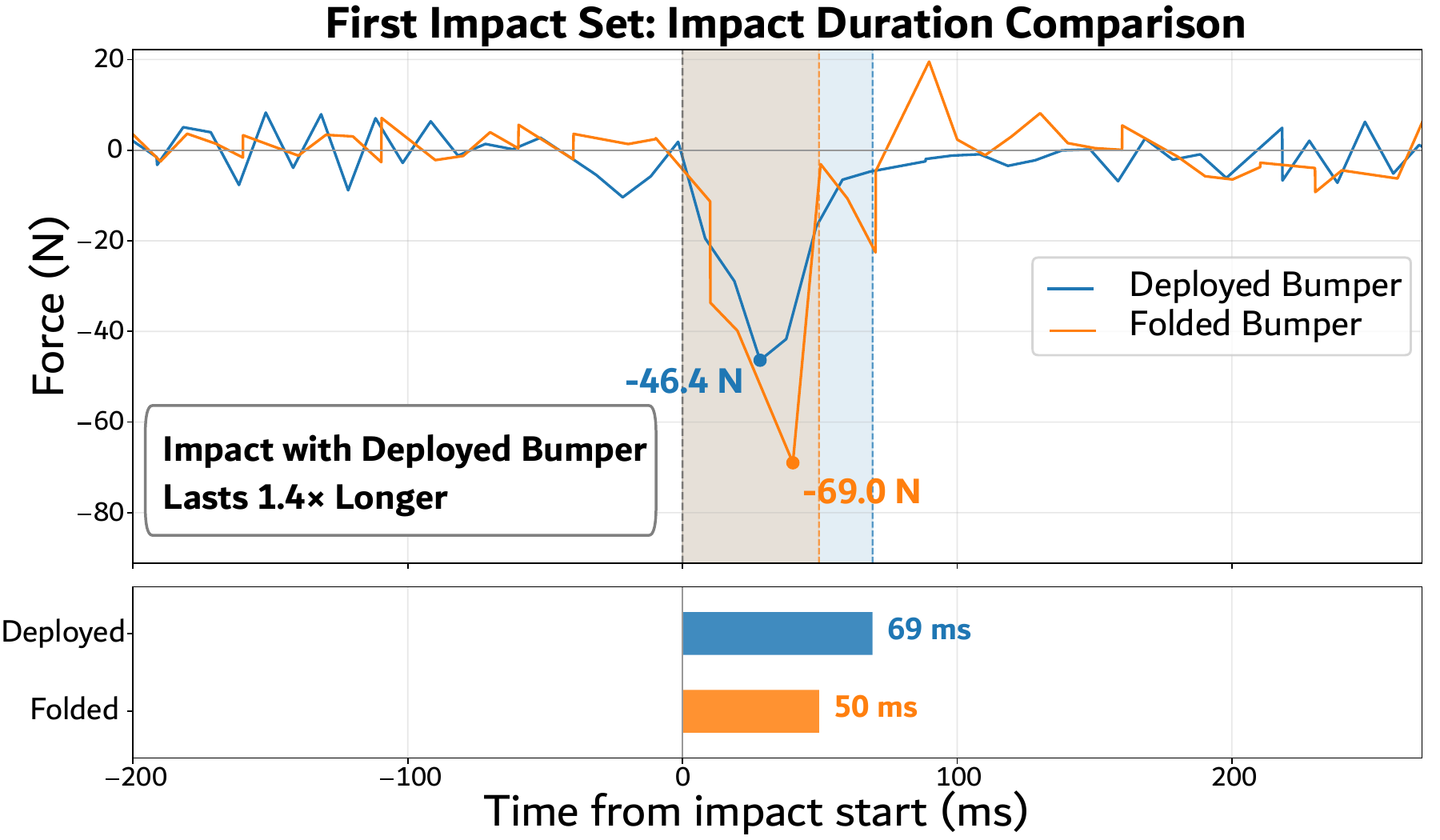}
  \caption{Impact duration}\label{fig:force-b}
\end{subfigure}
\caption{First impact set: the deployed exosuit lowers the peak reaction force
(a) and broadens the contact window (b), evidencing better energy dissipation.}
\label{fig:force}
\end{figure}
*/

This effect becomes even more pronounced when the compared impacts differ slightly in speed. In a second selected impact set, a modest increase in pre-impact speed of $0.3$~m/s (approximately $30\%$) for the folded configuration results in a disproportionate rise in peak acceleration, from $15.7$ to $103.9$~m/s$^2$, corresponding to an increase of $\sim$580\%. This disparity underlines the strong, nonlinear sensitivity of collision loads to impact velocity in the absence of a compliant protective structure, and highlights the risk of relying on rigid airframes alone in environments where impact speed cannot be tightly controlled. The measured metrics for both velocity-matched sets are summarized in Table~\ref{tab:results}. Considered together, across the two sets the deployed exoskeleton reduces peak reaction force by $33\%$ and $79\%$, respectively, and lengthens the impact duration by factors of $1.4\times$ and $1.9\times$ relative to the folded configuration. The correspondingly reduced jerk ($da/dt$) further implies gentler loading on the onboard IMU, electronics, and structural mounts, suggesting benefits for long-term component reliability that extend beyond immediate impact survivability.

Notably, the first deployed impact produced a combined reaction force of approximately $40$~N across the two bumpers without structural failure, exceeding the conservative $\approx20$~N static bound established by the FEA in Section~\ref{sec:fea}. This discrepancy is expected: a static analysis cannot capture the transient contact dynamics, strain-rate-dependent material response, and energy dissipation mechanisms present in a real collision, and therefore yields a conservative lower bound on load capacity rather than an accurate quantitative prediction of in-flight performance.

\begin{table}[h!]
\caption{Measured impact metrics (deployed vs.\ folded exoskeleton).}
\label{tab:results}
\centering\renewcommand{\arraystretch}{1.2}
\footnotesize
\begin{tabular}{l cc cc}
\toprule
 & \multicolumn{2}{c}{\textbf{Set 1}} & \multicolumn{2}{c}{\textbf{Set 2}} \\
\cmidrule(lr){2-3}\cmidrule(lr){4-5}
\textbf{Metric} & Depl. & Fold. & Depl. & Fold. \\
\midrule
Peak impact force (N)          & 46.4 & 69.0 & 20.4 & 96.7 \\
Max velocity (m/s)             & 1.3  & 1.3  & 1.1  & 1.6  \\
Peak acceleration (m/s$^2$)    & 35.7 & 53.1 & 15.7 & 103.9 \\
Impact duration (ms)           & 69   & 50   & 87   & 46   \\
Duration ratio (D$\div$F)      & \multicolumn{2}{c}{$1.4\times$} & \multicolumn{2}{c}{$1.9\times$} \\
\bottomrule
\end{tabular}
\end{table}

\subsection{Discussion}

While the experimental results validate the overall collision-tolerance strategy, several limitations of the current implementation define its operational envelope and motivate future work. 
The primary constraint is perceptual coverage: a single front-mounted RGB camera reliably supports deployment decisions for obstacles ahead of the UAV but leaves lateral and rear approaches undetected. This reflects a limitation of sensor placement rather than the language-driven perception model itself, indicating that omnidirectional protection will require complementary sensing (e.g., additional cameras or depth/ultrasonic sensors) rather than algorithmic refinement alone. 
A related limitation is the bumper's angular coverage, which mitigates impacts within a defined range around the airframe but not from directly above or below, motivating future geometries (e.g., curved or partially spherical designs) that better match the omnidirectional nature of potential threats. 
A further lesson concerns end-to-end latency: remote inference combined with a two-frame confirmation step to suppress false positives yields an approximate two-second delay between detection and effective deployment, restricting validated operation to low-to-moderate speeds. This suggests that the system's reaction-time budget, rather than the bumper's mechanical deployment speed, is the binding constraint on extending protection to faster flight. 
Finally, although the design integrates soft and rigid elements, deployment still relies on conventional rigid actuation. Compliant alternatives (e.g., pneumatic, tendon-driven, or shape-memory-based actuation) could reduce structural rigidity while preserving rapid, reliable deployment, further aligning the system with soft-robotics principles.

Collectively, these findings indicate that the main barriers to fully autonomous, high-speed, omnidirectional protection lie in sensing coverage, protective geometry, and decision latency, rather than in the core deployment concept, clarifying the engineering priorities for future iterations of the platform. Addressing these barriers represents a significant opportunity for the field, as it would enable UAVs to operate safely across a broader range of environments, both for the platforms themselves and for the humans in their vicinity. Continued investigation into strategies that enhance human--robot collaborative safety, while simultaneously improving robustness against both incidental and intentional collisions, therefore remains a compelling and consequential direction for future research.

\section{CONCLUSIONS}
This paper presented the design, control, and experimental evaluation of a soft, reconfigurable, shock-absorbing exoskeleton for a quadrotor platform. By retaining a standard rigid airframe and augmenting it with an origami-inspired TPU bumper that deploys only when required, the proposed system behaves as a compliant spring--damper during collision events while preserving nominal flight performance during unobstructed operation. A nonlinear hyperelastic finite element analysis guided the structural design and sizing of the bumper, while a CLIPSeg-based, language-driven perception service enabled autonomous, threshold-based deployment using a single onboard RGB camera, without requiring task-specific retraining of the underlying vision model. Motion-capture experiments conducted over matched-velocity impacts demonstrated that deploying the exoskeleton reduces peak impact force by up to $79\%$, lowers peak acceleration, and lengthens the contact duration by a factor of $1.4$--$1.9\times$ relative to the folded configuration, yielding smoother deceleration profiles and substantially reduced mechanical loads on the airframe. Taken together, these results constitute a complete proof of concept for a soft, reconfigurable collision-tolerance strategy for aerial robots, validated through both simulation and real-world flight testing, and establish a foundation for future work on closed-loop actuation, multi-directional protection, and broader obstacle taxonomies.


\section*{ACKNOWLEDGMENTS}
During the preparation of this work, the author(s) used the Claude AI-tool developed by Anthropic, in order to correct grammar and refine the wording of the manuscript.

\bibliographystyle{ieeetr}
\bibliography{references}

\end{document}